\documentclass{article}

\usepackage[final]{neurips_2025}
\usepackage{graphicx}
\usepackage{booktabs}
\usepackage{soul}
\usepackage{float}

\usepackage[utf8]{inputenc} % allow utf-8 input
\usepackage[T1]{fontenc}    % use 8-bit T1 fonts
\usepackage{hyperref}       % hyperlinks
\usepackage{url}            % simple URL typesetting
\usepackage{booktabs}       % professional-quality tables
\usepackage{amsfonts}       % blackboard math symbols
\usepackage{nicefrac}       % compact symbols for 1/2, etc.
\usepackage{microtype}      % microtypography
\usepackage{xcolor}         % colors
\usepackage[ruled,vlined]{algorithm2e}
\usepackage{amsmath,amssymb}

\definecolor{mygreen}{rgb}{0.13, 0.55, 0.13}

\makeatletter
\def\blfootnote{\xdef\@thefnmark{}\@footnotetext}
\makeatother

\title{Progressive Data Dropout: An Embarrassingly Simple Approach to Train Faster}

\author{%
  Shriram M S\textsuperscript{1*},
  Xinyue Hao\textsuperscript{2*},
  Shihao Hou\textsuperscript{3},
  Yang Lu\textsuperscript{3},
  Laura Sevilla-Lara\textsuperscript{2}, \\
  \textbf{Anurag Arnab}, %\textsuperscript{4},
  \textbf{Shreyank N Gowda\textsuperscript{4}} \\
  \textsuperscript{1}Department of Computer Science, University of Manchester \\
  \textsuperscript{2}School of Informatics, University of Edinburgh \\
  \textsuperscript{3}School of Informatics, Xiamen University \\
  \textsuperscript{4}School of Computer Science, University of Nottingham
}

\begin{document}

\maketitle

\begin{abstract}
The success of the machine learning field has reliably depended on training on large datasets. While effective, this trend comes at an extraordinary cost. This is due to two deeply intertwined factors: the size of models and the size of datasets. While promising research efforts focus on reducing the size of models, the other half of the equation remains fairly mysterious. Indeed, it is surprising that the standard approach to training remains to iterate over and over, uniformly sampling the training dataset. In this paper we explore a series of alternative training paradigms that leverage insights from hard-data-mining and dropout, simple enough to implement and use that can become the new training standard. The proposed Progressive Data Dropout reduces the number of effective epochs to as little as 12.4\% of the baseline. This savings actually do not come at any cost for accuracy. Surprisingly, the proposed method improves accuracy by up to 4.82\%. Our approach requires no changes to model architecture or optimizer, and can be applied across standard training pipelines, thus posing an excellent opportunity for wide adoption. Code can be found here: \url{https://github.com/bazyagami/LearningWithRevision}.
\end{abstract}

% \begin{abstract}
% Training deep neural networks is computationally expensive, often requiring significant time and energy \ls{power?} to converge on large datasets. In this paper, we propose Progressive Data Dropout, an embarrassingly simple yet effective approach to improve training efficiency by progressively discarding subsets of data across epochs. We explore three variants: (1) a curriculum-inspired method that initially focuses on the hard examples misclassified or predicted with low confidence; (2) a scalar-based \ls{I'd just say decay} decay method that randomly drops a fixed proportion of data each epoch; and (3) a hybrid approach that mimics the schedule of the first method but drops data at random rather than based on difficulty. All three approaches use the full training dataset only in the last epochs. Despite its simplicity, the third variant achieves the best performance in our experiments. Remarkably, our approach reduces the number of effective epochs to as little as 12.4\% of the baseline measured in `Effective Epochs', a hardware-independent proxy for backpropagation effort while improving accuracy by up to 4.82\%, depending on the model and dataset. Our approach requires no changes to model architecture or optimizer, and can be applied across standard training pipelines. These results demonstrate that carefully designed data dropout strategies can substantially reduce training costs while enhancing generalization. Code: \url{https://anonymous.4open.science/r/LearningWithRevision-1B1B}.
% \end{abstract}

\section{Introduction}
\label{sec:intro}
\blfootnote{* Equal Contribution.}
% Deep neural networks have revolutionized machine learning, achieving remarkable performance across diverse domains from computer vision \cite{he2016deep, dosovitskiy2020image} to natural language processing \cite{vaswani2017attention, brown2020language}. 
% However, these advances have often come at the cost of substantial computational resources, posing challenges for real-world applications, with unsustainable usage of electricity \cite{gowda2023watt}. 
Advances in deep learning over the last years have consistently come at the cost of substantial computational resources. Each year, the computational cost of the most notable models multiplies by over 4 times~\cite{epoch2023aitrends}. This poses challenges both in terms of real-world applications and sustainability~\cite{gowda2023watt}.

In response to these challenges, a significant body of research has emerged focused on developing efficient neural architectures. Models such as MobileNet \cite{howard2017mobilenets, sandler2018mobilenetv2}, EfficientNet \cite{tan2019efficientnet}, and EfficientFormer \cite{li2022efficientformer} have demonstrated that carefully designed architectures can achieve competitive performance with significantly reduced parameter counts and computational demands. These architectural innovations have been complemented by techniques such as pruning \cite{han2015learning}, quantization \cite{jacob2018quantization}, and knowledge distillation \cite{hinton2015distilling}, all aimed at reducing the resource footprint of deep neural networks.

While most efficiency efforts focus on the model, another underexplored direction is improving \emph{how} we train. Standard training treats all samples equally across all epochs, but not all data points are equally useful throughout the learning process. Some examples are consistently misclassified and carry high learning value, while others are confidently and correctly classified early on and may contribute little to further model improvement \cite{katharopoulos2018not}. Allocating equal computation to both types may be wasteful. Human learning strategies offer valuable insights for improving machine learning systems. When studying for examinations, humans rarely allocate equal attention to all material. Instead, they often adopt strategies that prioritize challenging concepts while ensuring comprehensive knowledge through periodic revision \cite{dunlosky2013improving, bjork2011making}. This selective focus on difficult content, followed by holistic review, allows for efficient allocation of cognitive resources while maintaining broad coverage of the subject matter. A parallel in deep learning is dropout \cite{srivastava2014dropout}, where neurons are randomly deactivated during training to prevent co-adaptation and improve generalization. However, dropout operates on the model's activations rather than the data. In contrast, data dropout~\cite{wang2018data}, the idea of selectively dropping training samples during learning, remains relatively unexplored.

In this paper, we introduce Progressive Data Dropout (PDD), a simple yet surprisingly effective family of strategies that progressively drops subsets of training data across epochs. Our methods are extremely easy to implement and grounded in both cognitive principles and practical efficiency goals. We explore three variants: a difficulty-based approach that initially trains only on hard examples; a scalar-based approach that randomly drops a fixed proportion of data each epoch; and a hybrid strategy that drops random samples using the same schedule as the difficulty-based method.

\begin{figure}[t]
    \centering
    \begin{minipage}{0.49\linewidth}
        \centering
        \includegraphics[width=\linewidth]{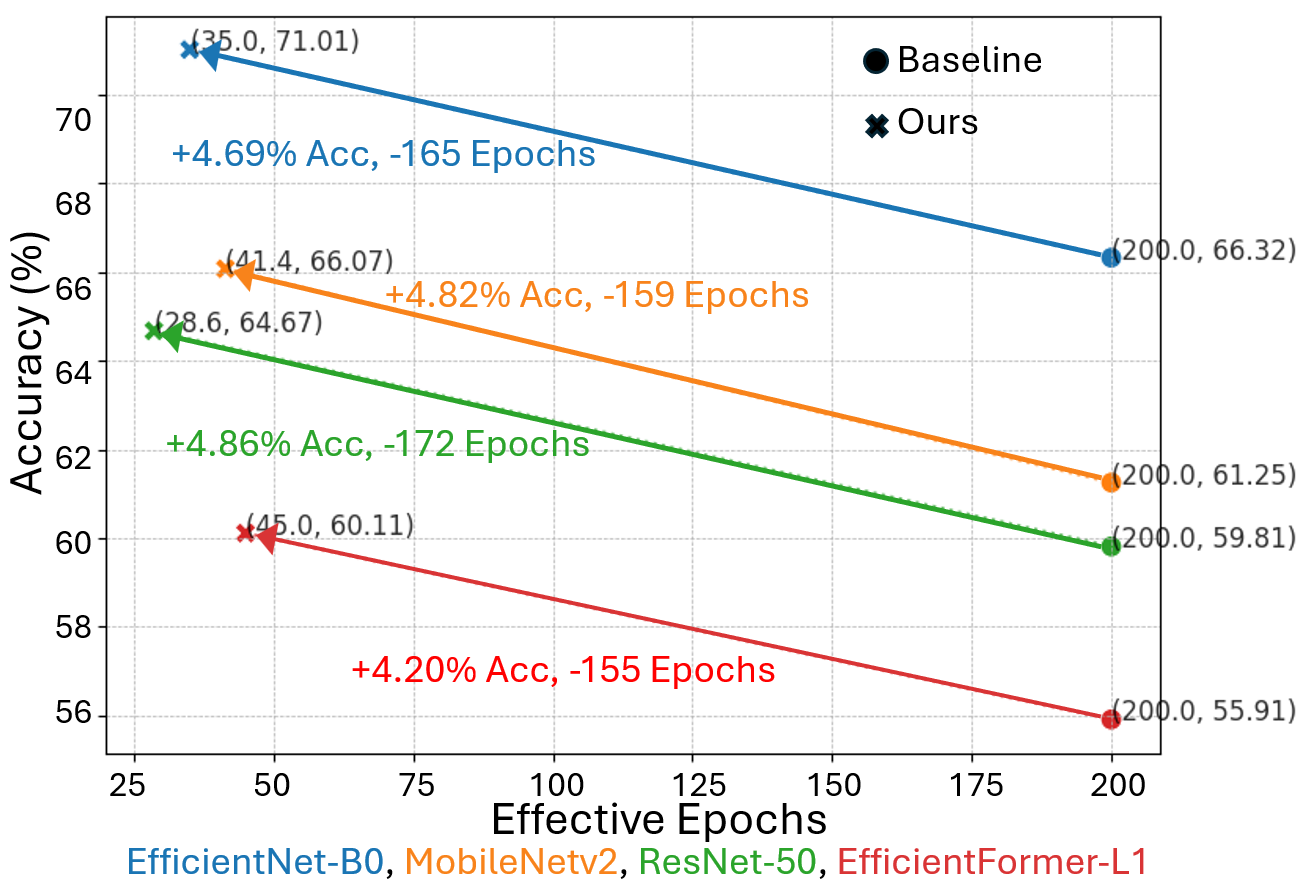}
        \caption{Using the proposed variants of Progressive Data Dropout, we can not only outperform the traditional baseline, but we can reduce the number of effective epochs significantly.This finding is consistent across a range of models and datasets.
        }
        \label{fig:teaser}
    \end{minipage}
    \hfill
    \begin{minipage}{0.49\linewidth}
        \centering
        \includegraphics[width=\linewidth]{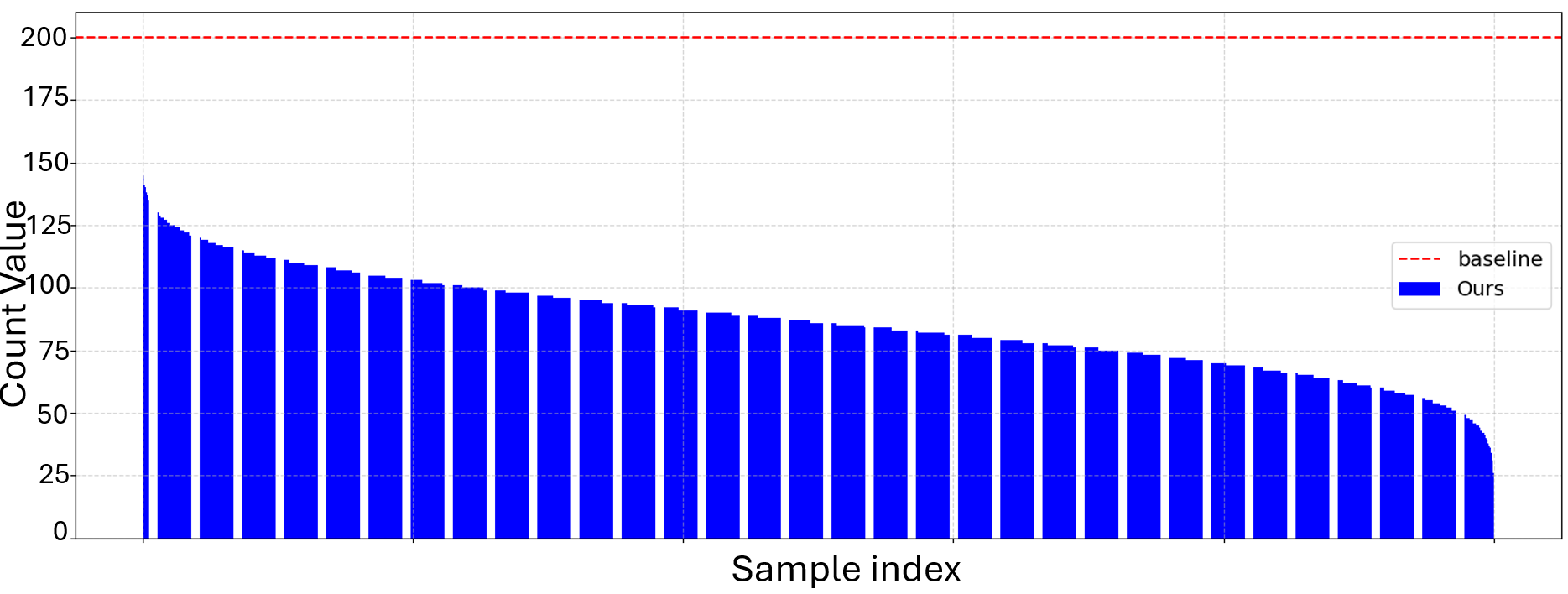}
        \caption{A possible reasoning for the effectiveness of our approach is that random dropout maintains representative coverage and introduces beneficial stochasticity. Here, we see an example of the number of times each sample goes through backpropagation. Traditionally, each sample would go through this `total epochs' number of times (red line).
        }
        \label{fig:randomselected}
    \end{minipage}
\end{figure}

Beyond the cognitive inspiration, our approach offers significant practical benefits. By initially focusing on difficult examples, the model processes substantially fewer data points during early epochs, leading to reduced computational requirements. This selective focus not only accelerates training but also appears to guide the model toward more generalizable representations, as evidenced by improved performance at test time. Our experimental results demonstrate that models trained with our approach achieve comparable or superior accuracy while requiring only \textbf{0.124$\times$} the number of effective epochs compared to standard training procedures. This efficiency gain is particularly notable for already-efficient architectures like MobileNet and EfficientFormer, where further reducing training costs can significantly impact deployment scenarios with constrained resources. We plot some of these results on CIFAR-100~\cite{krizhevsky2009learning} in Figure~\ref{fig:teaser}.

%\shreyank{}{Do I need to add something about why we are reporting number of training steps as a form of efficiency?.}\anurag{Just briefly say that its directly proportional to the total wallclock training time?}

Our work makes the following key contributions:

\begin{itemize}
    \item We propose \textit{Progressive Data Dropout}, a simple and general method to reduce training cost by progressively dropping subsets of data across epochs, inspired by cognitive learning strategies and model regularization techniques.
    
    \item We design and investigate three dropout variants: (1) a difficulty-based schedule that retains only hard examples early in training, (2) a scalar-based random sampling method, and (3) a schedule-matched random dropout strategy that surprisingly yields the best accuracy.
    
    \item Our approach significantly reduces effective epochs by up to 89.5\% fewer epochs compared to standard training, whilst still improving or matching accuracy, demonstrating effectiveness on multiple architectures including EfficientNet, MobileNet-V2, ResNet, EfficientFormer and ViT-B-MAE. We also show the potential cost saving in self-supervised learning.
    
    \item We show that Progressive Data Dropout is compatible with standard supervised training pipelines and can be easily integrated without architectural or optimization changes.
\end{itemize}

\section{Related Work}

Several learning paradigms share a similar underlying goal with our proposed method, namely, improving training efficiency and generalization, though they differ significantly in implementation and motivation. Curriculum learning, active learning, and data pruning are all relevant lines of research that intersect with our objectives. While we briefly discuss these in the below to contextualize our contributions, we focus our detailed comparisons on methods that explicitly implement forms of data dropout, given the page constraints.

\paragraph{Model Compression and Training Efficiency.}
Significant progress has been made in making neural networks more efficient at inference time through model compression techniques. Pruning \cite{han2015learning}, quantization \cite{jacob2018quantization}, and knowledge distillation \cite{hinton2015distilling} reduce the size or precision of model weights, allowing faster inference with minimal accuracy loss. Parallel efforts in efficient architecture design, such as MobileNet \cite{howard2017mobilenets}, EfficientNet \cite{tan2019efficientnet}, and EfficientFormer \cite{li2022efficientformer}, focus on reducing the number of parameters and FLOPs during inference. In contrast, our work targets the training phase rather than inference, aiming to reduce effective epochs and data usage dynamically, independent of the underlying model architecture.

\paragraph{Curriculum Learning and Sample Importance.} 
Curriculum learning \cite{bengio2009curriculum} and hard example mining \cite{shrivastava2016training} propose organizing training data to emphasize samples based on their difficulty or relevance. These methods aim to present samples in a structured order to accelerate convergence or improve performance. More recent work \cite{katharopoulos2018not} studies the informativeness of data during training and suggests adaptive data selection to optimize learning. Our work differs in that we progressively drop samples rather than reordering or reweighting them. Furthermore, we introduce both difficulty-based and random dropping strategies, showing that even unstructured data dropout can lead to improved generalization.

\paragraph{Training on Subsets and Data Pruning.}
%\anurag{I feel this is too vague. And these related works are quite related to our method. So I think that you should include citations in the following sentence for each point.}
Several recent works investigate training on a reduced subset of the data to minimize computational overhead. Data Diet \cite{tonevaempirical} proposes filtering data based on metrics like gradient norm and error L2-norm to identify examples that are likely to be forgotten during training. Dataset pruning methods, such as those proposed by Yang et al.~\cite{yang2022dataset}, aim to identify a minimal subset of training data that closely preserves the generalization performance of the full dataset, using optimization techniques with theoretical guarantees. Coleman et al.~\cite{colemanselection} introduce a ``selection via proxy'' (SVP) approach that accelerates data selection by training small proxy models instead of the full-scale target network, enabling faster core-set selection or active learning.  While related in goal, our method differs in important ways. Many of these approaches require substantial pre-processing or side-training before model training begins, for example, Data Diet requires tracking forgetting events early in training \cite{tonevaempirical}, and SVP uses proxy models trained in advance \cite{colemanselection}. Influence-based selection methods such as \cite{koh2017understanding, feldman2020neural} estimate the effect of removing samples on model predictions, which can be computationally expensive. Score-based heuristics (e.g., confidence scores or gradient magnitudes) also require repeated forward or backward passes over the entire training set \cite{mirzasoleiman2020coresets, mindermann2022prioritized}. In contrast, our approach is integrated directly into the training loop, progressively dropping examples over epochs without requiring separate scoring, proxy models, or influence estimation.

%\anurag{We also need to talk about active learning here -- Given a large unlabeleld dataset, active learning works out which samples to label and train on.}
\paragraph{Active Learning.} Active learning~\cite{settles2009active} aims to reduce labeling and training costs by selectively querying the most informative data points from an unlabeled pool. Common strategies include uncertainty sampling~\cite{nguyen2022measure}, where examples with the lowest prediction confidence are chosen; margin sampling~\cite{balcan2007margin}, which selects samples near the decision boundary; and query-by-committee~\cite{kee2018query}, which leverages disagreement among multiple models to identify uncertain data. These methods are typically iterative and assume access to large unlabeled datasets alongside a labeling budget.  In contrast, our approach operates in the fully supervised regime and does not require querying new data, rather, we progressively drop portions of the labeled dataset during training to reduce computation and potentially improve generalization.

\paragraph{Data Dropout and Data-Centric Regularization.}
Dropout \cite{srivastava2014dropout} is a widely-used regularization method that randomly drops activations during training to prevent co-adaptation and overfitting. This concept has inspired architectural variants such as DropConnect \cite{wan2013regularization} and stochastic depth \cite{huang2016deep}. While these methods manipulate the network topology, relatively fewer studies have explored analogs at the data level. 
Most closely related to our method are approaches that drop data samples during training in order to improve efficiency or model generalization. These \textit{data dropout} techniques differ in motivation and implementation, ranging from data augmentation to importance-based filtering. Some methods focus on data-level perturbations for augmentation purposes. For instance, Generalized Dropout \cite{rahmani2018data} applies dropout directly to image pixels, generating additional variations of the same training sample to improve generalization. While effective as a form of regularization, these approaches do not aim to reduce the number of training iterations or epochs. Other works emphasize importance-based data pruning. Katharopoulos and Fleuret \cite{katharopoulos2018not} propose a sampling strategy that prioritizes high-gradient or high-loss examples, dropping low-impact samples to accelerate convergence. Wang et al.~\cite{wang2018data} introduce a two-round training scheme in which training samples are evaluated after an initial round and permanently dropped if they are deemed to negatively influence generalization. Similarly, DropSample \cite{yang2016dropsample} and Greedy DropSample \cite{yang2020accelerating} propose sample selection mechanisms that adjust data retention dynamically, particularly in the context of Chinese character classification. Zhong et al.~\cite{zhong2021dynamic} propose Dynamic Training Data Dropout (DTDD), which filters noisy samples over several epochs to improve robustness in deep face recognition. In contrast to all these methods, our approach is integrated directly into the training loop and operates in a progressive manner. Rather than relying on pretraining, expensive influence function estimation (which require second-order gradients), or permanently dropping samples after a single evaluation, we iteratively modify the training set across epochs using cognitively inspired and stochastic strategies. This allows us to balance efficiency and accuracy without the need for auxiliary models, validation sets, or expensive influence computation.

\section{Method}

%\anurag{We can maybe include an Algorithm / pseudocode here, to make it clear how easy this is to implement.}

We introduce \textit{Progressive Data Dropout}, an embarrassingly simple yet effective method to accelerate neural network training by progressively reducing the training set across epochs. Inspired by cognitive science and dropout regularization, our approach incrementally drops training samples based on either difficulty or randomized criteria, while ensuring full-data coverage in the final epoch. In this section, we describe three variants of our method. A full algorithm explaining the approach can be found in the supplementary material.

\subsection{Overview}

Let $\mathcal{D}_0 = \{(x_i, y_i)\}_{i=1}^{N}$ be the original training dataset. At each epoch $t$, we construct a subset $\mathcal{D}_t \subseteq \mathcal{D}_0$ such that fewer samples are used as training progresses. This reduction can be based either on model confidence (difficulty-based) or through random sampling. In all variants, the final epoch uses the full dataset $\mathcal{D}_0$, mimicking the ``revision'' phase in human learning where comprehensive review reinforces generalization.

%Let $\mathcal{D}_0 = \{(x_i, y_i)\}_{i=1}^{N}$ be the original training dataset. At each epoch $t$, we construct a subset $\mathcal{D}_t \subseteq \mathcal{D}_0$ such that fewer samples are used in earlier epochs \anurag{provide some intuition for this}, and more are used as training progresses. In all variants, the final epoch is performed on the full dataset $\mathcal{D}_0$, mimicking the ``revision'' phase in human learning.

\subsection{Variant 1: Difficulty-First Training}

This cognitively motivated strategy~\cite{bjork2011making} focuses on ``hard'' examples early on and uses all examples at the last epoch as a form of `revision'. After each epoch, we evaluate the model’s confidence or correctness on each sample:

\begin{itemize}
    \item Let $p_t(x_i)$ be the model’s predicted probability for the correct class of sample $x_i$ at epoch $t$.
    \item A sample is considered ``easy'' if $p_t(x_i) \geq \tau_t$, where $\tau \in [0,1]$ is a fixed threshold .
    \item The dataset for the each epoch is:
    %\anurag{The dataset is getting smaller over time, which is the opposite of what the intro paragraph says about the dataset getting larger over time}:
    \[
    \mathcal{D}_{t} = \{(x_i, y_i) \in \mathcal{D} \mid p_t(x_i) < \tau \} \; \forall t=0,..,T
    \]
    \item In the final epoch, we reset to $\mathcal{D}_T = \mathcal{D}_0$.
\end{itemize}

This approach mimics human study habits: initially focusing on difficult material, then performing a complete review. We refer to this method as \textbf{Difficulty-Based Progressive Dropout} (DBPD).

\subsection{Variant 2: Scalar-Based Random Dropout}

This variant eschews difficulty estimation and instead uses a scalar decay factor $\alpha \in (0, 1)$ to progressively reduce the dataset:

\begin{itemize}
    \item The number of samples in epoch $t$ is $|\mathcal{D}_t| = \alpha^t \cdot N$.
    \item $\mathcal{D}_t$ is created by randomly sampling (without replacement) from $\mathcal{D}_0$.
    \item The full dataset is used in the final epoch: $\mathcal{D}_T = \mathcal{D}_0$.
\end{itemize}

This method, which we call \textbf{Scalar Random Dropout} (SRD), is the simplest to implement and requires no per-sample evaluation. Despite its simplicity, it yields surprisingly competitive performance with substantial computational savings.

\subsection{Is Difficulty Scoring Necessary? Variant 3: Schedule-Matched Random Dropout}

To test whether explicit difficulty-based scoring is essential for effective data dropout, we design a variant that uses the same per-epoch data reduction schedule as Variant 1 (difficulty-based), but drops samples \textit{at random} rather than based on confidence.

\begin{itemize}
    \item Let $|\mathcal{D}_t^{(1)}|$ be the dataset size at epoch $t$ under the difficulty-based method.
    \item In this variant, we randomly subsample $|\mathcal{D}_t^{(1)}|$ points from $\mathcal{D}_0$ at each epoch $t$, without regard to difficulty.
    \item As in the other methods, the final epoch uses the full dataset: $\mathcal{D}_T = \mathcal{D}_0$.
\end{itemize}

We refer to this as \textbf{Schedule-Matched Random Dropout} (SMRD). While this approach is less practical to implement in real-time (as it relies on the difficulty-based schedule computed in a separate run), it serves as a valuable ablation to probe the importance of difficulty estimation. Surprisingly, SMRD achieves the highest test accuracy across our experiments, indicating that structured data reduction alone, without any notion of sample difficulty, can yield strong generalization benefits.

To approximate the schedule without explicitly scoring sample difficulty, we also explore a mathematical approximation of the number of samples dropped per epoch. Results of this approximation are included in the supplementary material.

\subsection{Effective Epochs}

To provide a hardware-independent measure of training efficiency, we introduce the notion of \textit{Effective Epochs} (EE). While traditional epoch counts assume that every training sample undergoes both forward and backward passes, our method performs a forward pass on all samples but selectively applies backpropagation only to a subset. This design reduces computational cost while retaining compatibility with standard data pipelines.

We define the number of \textit{Effective Epochs} as:
\[
\text{Effective Epochs} = \frac{\text{Number of Samples through Backpropogation over entire run}}{\text{Total Number of Training Samples}}
\]

This quantity reflects how many full passes over the dataset (with backpropagation) were effectively executed. It captures the true training burden independent of batch size, hardware, or epoch length. Throughout the paper, we report effective epochs as our standard efficiency metric to quantify the computational savings achieved by our dropout strategies.

\section{Discussion}

Progressive Data Dropout (PDD) improves generalization by gradually reducing the training dataset during learning. At each epoch, a portion of data is dropped, randomly or based on difficulty, forcing the model to focus on more informative examples and avoid overfitting. This acts as a form of data-level regularization, analogous to dropout on model activations~\cite{srivastava2014dropout}. Surprisingly, random dropout outperforms difficulty-based dropout, suggesting that the benefit comes more from data reduction than selective removal. However, just randomly dropping a percentage of data does not work best (see SRD). However, having a scheduled based random dropout works best. Whilst, we try to mathematically approximate this and test, we do not obtain our best results, leaving this to future work.  Difficulty-based methods may skew the training distribution or overfit to noisy samples \cite{yang2016dropsample, yang2020accelerating}, while random dropout maintains representative coverage and introduces beneficial stochasticity.
%\anurag{Also the fact that once you drop, you never see again.}
PDD builds on insights from dataset pruning and curriculum learning. Prior work shows that large portions of training data can be discarded with minimal or even positive effects on accuracy \cite{tonevaempirical, yang2022dataset}. Unlike methods requiring retraining, proxies, or selection phases \cite{wang2018data, colemanselection}, PDD dynamically reduces data within a single run.
Unlike curriculum learning that adds or reweights samples \cite{bengio2009curriculum, shrivastava2016training}, PDD follows an anti-curriculum, reducing data as training progresses. Its success, especially in the random variant, highlights that training with less but diverse data can improve robustness and generalization.

\section{Experimental Analysis}

\subsection{Datasets}

We conduct all classification experiments using three standard image classification benchmarks: CIFAR-10~\cite{krizhevsky2009learning}, CIFAR-100~\cite{krizhevsky2009learning}, and ImageNet~\cite{deng2009imagenet}. The CIFAR-10 and CIFAR-100 datasets each consist of $60{,}000$ $32 \times 32$ color images, split into $50{,}000$ training and $10{,}000$ test images. CIFAR-10 contains 10 object classes, while CIFAR-100 spans a more fine-grained taxonomy of 100 classes. Both datasets are widely used to benchmark small and mid-sized neural architectures. We also use the ImageNet (ILSVRC2012) dataset, which contains over $1.28$ million training images and $50{,}000$ validation images across $1{,}000$ object categories. The images are of higher resolution and greater diversity compared to CIFAR datasets, making ImageNet a robust benchmark for evaluating both model capacity and training efficiency.
In addition to supervised classification, we also evaluate our method in a self-supervised learning setup, using ImageNet as the benchmark dataset. This allows us to assess the generality of Progressive Data Dropout beyond supervised learning scenarios.

%\anurag{Also need to consider training-from-scratch, vs finetuning}

\subsection{Implementation Details}

All variants are implemented with minimal overhead on top of standard training loops. To avoid introducing instability, batch normalization statistics and learning rates are kept consistent with standard training procedures. For variants with data reduction, we ensure that the data loader is reset at each epoch to reflect the newly selected subset. Importantly, we do \textbf{not} alter model architectures, loss functions, or optimizers. Our method is architecture-agnostic and seamlessly integrates with any supervised training pipeline. We use popular models such as MobileNetv2~\cite{sandler2018mobilenetv2}, EfficientNet~\cite{tan2019efficientnet}, Efficientformer~\cite{li2022efficientformer}, ResNet~\cite{he2016deep} and ViT-B~\cite{dosovitskiy2020image} for the image classification tasks. We also consider MAE~\cite{he2022masked} and perform masked pre-training using the Scalar Dropout idea and fine-tune the model using standard fine-tuning. We follow the official implementation details for ImageNet experiments, so for example EfficientNet is run for 350 epochs, while ResNet-50 is run for 100 epochs. However, for all CIFAR-100 experiments we run all models for 200 epochs and for all CIFAR-10 experiments we run them for 30 epochs. For our CIFAR experiments, we use an AdamW optimizer with a StepLR scheduler with a step size of 1. Our initial learning rate is $0.0003$. For the self-supervised learning experiments with MAE~\cite{he2022masked}, we follow standard practice~\cite{he2016deep} and run the pre-training phase for 800 epochs. We use a fixed batch size of 32 for all our experiments. We use 8 A100 GPUs with 40GB memory. For ImageNet fine-tuning we use 1 A100 GPU with 40GB memory. For CIFAR10 and CIFAR100, we use a 4060RTX GPU with 8GB memory.

\subsection{Supervised Image Classification}

In Table~\ref{tab:cifar}, we compare the performance of EfficientNet-B0~\cite{tan2019efficientnet}, MobileNet-V2~\cite{sandler2018mobilenetv2}, ResNet-50~\cite{he2016deep}, EfficientFormer~\cite{li2022efficientformer}, and ViT-B-MAE~\cite{he2022masked} across CIFAR-10, CIFAR-100, and ImageNet. We report results for each model under the baseline training regime as well as the three main variants of our proposed method: DBPD 0.3, SMRD 0.3, and SRD 0.95. For completeness, results for DBPD 0.7 and SMRD 0.7 are provided in the supplementary material. Among all configurations, SMRD 0.3 consistently achieves the highest accuracy across most settings, while the 0.3 threshold variants also offer the greatest reduction in effective epochs. Across datasets and architectures, our approach yields improvements in accuracy of up to \textbf{4.82}\% and reduces the effective epochs by as much as \textbf{12.4}\%.

\begin{table}
\caption{Accuracy (\%) and effective epochs reported as \textit{accuracy/effective epochs}. All methods besides ViT-MAE are trained from scratch unless otherwise mentioned. ViT-MAE uses the publicly available checkpoint~\cite{he2022masked}. All results have been conducted following official implementations on timm~\cite{timm}. Best results in \textcolor{blue}{blue} and second best in \textcolor{mygreen}{green}. The last column shows the gains in accuracy and drop in effective epochs of the SMRD variant.
}
\resizebox{\linewidth}{!}{
\begin{tabular}{l cccc cc}
\midrule
                       & Baseline                  & DBPD & SRD & SMRD & Acc. gain (\%) & EE saved \\ \midrule
\multicolumn{1}{l}{\textit{CIFAR10}}     &  &   &  &  &     &         \\ 
\multicolumn{1}{l}{EfficientNet-B0} & 90.41 / 200 & 88.22 / \textcolor{blue}{18} & \textcolor{blue}{92.07} / 84 & \textcolor{mygreen}{90.55} / \textcolor{mygreen}{20} & \textit{0.14} & \textit{10} $\times$\\
\multicolumn{1}{l}{MobileNet-V2}   & 88.46 / 200 & 89.55 / \textcolor{blue}{22.3} & \textcolor{mygreen}{90.73} / 84 & \textcolor{blue}{90.91} / \textcolor{mygreen}{24.0} & \textit{2.45}  & \textit{8.33} $\times$       \\
ResNet-50 & 87.94 / 200 & 88.42 / \textcolor{mygreen}{23}  & \textcolor{blue}{89.81} / 84 & \textcolor{mygreen}{89.12} / \textcolor{blue}{22.1} & \textit{1.18} & \textit{9.05} $\times$ \\ 
EfficientFormer-L1 & 82.38 / 200 & \textcolor{mygreen}{84.47} / \textcolor{blue}{28.2} & 84.38 / 84 & \textcolor{blue}{85.20} / \textcolor{mygreen}{31} & \textit{2.82} &  \textit{6.45} $\times$ \\ \midrule
\multicolumn{1}{l}{\textit{CIFAR100}}     &  &   &  &  &     &     \\ 
\multicolumn{1}{l}{EfficientNet-B0} & 66.32 / 200 & \textcolor{mygreen}{67.15} / \textcolor{blue}{24.8}& 66.75 / \textcolor{blue}{24.8} & \textcolor{blue}{71.01} / \textcolor{mygreen}{35} & \textit{4.69} & \textit{5.71} $\times$ \\
\multicolumn{1}{l}{MobileNet-V2} &    61.25 / 200 & \textcolor{mygreen}{62.85} / \textcolor{mygreen}{29.6} & 62.80 / \textcolor{blue}{24.8} & \textcolor{blue}{66.07} / 41.4 & \textit{4.85} & \textit{4.83} $\times$\\
ResNet-50       & 59.81 / 200 & 60.13 / \textcolor{blue}{16.26} & \textcolor{mygreen}{60.83} / \textcolor{mygreen}{24.8} & \textcolor{blue}{64.67} / 28.6 & \textit{4.86} & \textit{6.99} $\times$ \\ 
EfficientFormer-L1  & 55.91 / 200 & \textcolor{mygreen}{57.62} / \textcolor{mygreen}{30} & 56.95 / \textcolor{blue}{24.8} & \textcolor{blue}{60.11} / 45 & \textit{4.20} & \textit{4.44} $\times$ \\
\midrule
\multicolumn{4}{l}{\textit{CIFAR100 finetuned from ImageNet}}     &  &    &         \\ 
\multicolumn{1}{l}{EfficientNet-B0} & 83.31 / 200 & \textcolor{mygreen}{83.95} / \textcolor{mygreen}{25.4} & 83.58 / \textcolor{blue}{24.8} & \textcolor{blue}{84.45} / 33.8 & \textit{1.14} & \textit{5.92} $\times$\\
\multicolumn{1}{l}{MobileNet-V2}    & 74.10 / 200 & \textcolor{mygreen}{74.35} / \textcolor{blue}{22.4} & 74.22 / \textcolor{mygreen}{24.8} & \textcolor{blue}{74.73} / 36.2 & \textit{0.63} & \textit{5.52} $\times$ \\
ResNet-50 & 78.59 / 200 & \textcolor{mygreen}{79.92} / \textcolor{blue}{21} & 79.63 / \textcolor{mygreen}{24.8} & \textcolor{blue}{81.54} / 33.2 & \textit{2.95} & \textit{6.02} $\times$ \\ 
EfficientFormer-L1 & 85.30 / 200 & \textcolor{mygreen}{86.74} / \textcolor{mygreen}{36.6} & 84.78 / \textcolor{blue}{24.8} & \textcolor{blue}{87.51} / 51.2 & \textit{2.21} & \textit{3.91} $\times$\\
ViT-B-MAE  & \textcolor{mygreen}{86.81} / 200 & 85.92 / \textcolor{mygreen}{52.2} & 82.90 / \textcolor{blue}{24.8} & \textcolor{blue}{88.10} / 67.2 & \textit{1.29} & \textit{2.98} $\times$\\
\midrule
\multicolumn{1}{l}{\textit{ImageNet}}     &  &   &  &  &     &      \\ 
EfficientNet-B0 & 77.10 / 350 & \textcolor{mygreen}{77.45} / \textcolor{mygreen}{111.2} & 64.71 / \textcolor{blue}{20} & \textcolor{blue}{79.51} / 142.4 & \textit{2.41} & \textit{2.46} $\times$ \\
MobileNet-V2 & \textcolor{mygreen}{71.60} / 250 & 68.42 / \textcolor{mygreen}{54.5} & 63.85 / \textcolor{blue}{20} & \textcolor{blue}{73.25} / 98.0 & \textit{1.75} & \textit{2.55} $\times$ \\
ResNet-50 & 75.04 / 100 & \textcolor{mygreen}{76.21} / \textcolor{mygreen}{33.0} & 71.54 / \textcolor{blue}{20} & \textcolor{blue}{79.18} / 46.4 & \textit{4.14} & \textit{2.15} $\times$ \\ 
EfficientFormer-L1 & 79.11 / 300 & \textcolor{mygreen}{79.24} / \textcolor{mygreen}{91.5} & 69.67 / \textcolor{blue}{20} & \textcolor{blue}{80.25} / 106.1 & \textit{1.14} & \textit{2.83} $\times$ \\
ViT-B-MAE & 83.10 / 100 & \textcolor{mygreen}{83.31} / \textcolor{mygreen}{35.4} & 79.51 / \textcolor{blue}{20} & \textcolor{blue}{84.13} / 41.7 & \textit{1.03} & \textit{2.40} $\times$ \\
\bottomrule

\end{tabular}}
\label{tab:cifar}
\end{table}

\subsection{Self-Supervised Training}

Self-supervised learning is among the most computationally expensive training paradigms, often accounting for over 95\% of the total training cost when including fine-tuning across multiple downstream tasks. Masked Autoencoders (MAE)~\cite{he2022masked} have shown strong promise in learning generalizable representations and improving performance on a wide range of tasks. In this work, we focus on the ViT-MAE-B model. While the original implementation pre-trains this model for 1600 epochs, we adopt the 800-epoch variant due to memory and compute limitations. Notably, the drop in accuracy for the 800-epoch model on ImageNet is less than 0.3\%, making it a reasonable compromise.

To apply Progressive Data Dropout in this setting, we use the SRD 0.98 variant, which is the simplest to adapt to a self-supervised objective. Additional results using other variants are included in the supplementary material. Following pretraining, we fine-tune the model using the official MAE procedure on ImageNet and CIFAR-100, and also evaluate it on the COCO object detection benchmark to test cross-task generalizability.

Despite its simplicity, SRD reduces the total number of effective epochs from 800 to just 50, a \textbf{16$\times$ reduction} in training steps. This reduction is especially impactful given the extreme time, memory, and energy requirements of MAE pretraining. Moreover, across all downstream tasks, our method remains highly competitive with the baseline, with only negligible loss in performance. We report these results in Table~\ref{tab:mae_srd_comparison}.

\begin{table}[t]
\centering
\begin{minipage}{0.49\linewidth}
\centering
\caption{Comparison of effective epochs and accuracy for ViT-MAE-B under standard MAE (baseline) and SRD 0.98 configurations across various tasks. Accuracy is reported as Top-1 (\%) for classification tasks and mAP (\%) for object detection.}
\resizebox{\linewidth}{!}{
\begin{tabular}{l|c|c|c|c}
\toprule
\textbf{Model} & \textbf{Pretraining} & \textbf{ImageNet} & \textbf{CIFAR-100} & \textbf{COCO} \\
 & EE & Accuracy & Accuracy & AP$^{box}$ \\
\midrule
Baseline & 800 & 83.10\% & 86.81\% & 49.9 \\
SRD 0.98 & \textbf{49.95} & 77.89\% & 83.81\% & 47.2 \\
\bottomrule
\end{tabular}
}
\label{tab:mae_srd_comparison}
\end{minipage}
\hfill
\begin{minipage}{0.49\linewidth}
\centering
\caption{Baseline trained to match the effective number of epochs (EE) used by dropout variants. }
\resizebox{\linewidth}{!}{
\begin{tabular}{l|c|c|c}
\toprule
\textbf{Variant} & \textbf{EE} & \textbf{Baseline (\%)} & \textbf{Accuracy (\%)} \\
\midrule
DBPD 0.3 & 25 & 63.32 & 67.15 \\
DBPD 0.7 & 33 & 63.57 & 68.80 \\
SRD 0.95 & 25 & 63.32 & 66.75 \\
SMRD 0.3 & 35 &63.52 & 71.01 \\
SMRD 0.7 & 46 &63.88 & 70.75 \\
\bottomrule
\end{tabular}
}
\label{tab:matched_baseline}
\end{minipage}
\end{table}

%\subsection{Comparison To Other Data Dropout Approaches}

%\anurag{For the ablation experiments, its also important to add an experiment that breaks things. As in, if we don't do X, then the performance becomes really bad. That will give the reader a much better understanding of why the method works.}

\subsection{Comparison to State-of-the-Art}
We conducted comparisons against a suite of existing methods, including DataDiet (ELN and Forget)~\cite{paul2021deep}, IES~\cite{yuan2025instancedependent}, Early Stopping, and InfoBatch~\cite{qininfobatch}. We followed the official implementations for all baselines and used EE to evaluate efficiency. Our method outperforms all compared approaches in both accuracy and EE, with higher values indicating better performance for both metrics. The results, reported as Accuracy / EE, are obtained by fine-tuning models on CIFAR-100 (pretrained on ImageNet) for 200 epochs.

\begin{table}[h]
\centering
\caption{Comparison with strong active learning baselines on CIFAR-100.}
\label{tab:active_learning_cifar100}
\resizebox{\linewidth}{!}{
\begin{tabular}{l|c|c|c|c|c|c|c}
\toprule
\textbf{Model} & \textbf{Baseline} & \textbf{DataDiet ELN} & \textbf{DataDiet Forget} & \textbf{IES} & \textbf{Early Stopping} & \textbf{InfoBatch} & \textbf{SMRD (ours)} \\
\midrule
ResNet-50 &  78.5 / 200  &  78.6 / 141  &  79.1 / 142  &  79.5 / 157  &  74.3 / 121  &  80.4 / 145  &  \textbf{81.5 / 33}  \\
EfficientNet &  83.3 / 200  &  82.7 / 139  &  83.0 / 139  &  83.7 / 153  &  76.6 / 132  &  84.0 / 141  &  \textbf{84.4 / 34}  \\
EfficientFormer-L1 &  85.3 / 200  &  84.8 / 142  &  85.0 / 140  &  85.8 / 155  &  78.2 / 125  &  86.6 / 143  &  \textbf{87.5 / 51}  \\
\bottomrule
\end{tabular}}
\end{table}

\subsection{Generalization in Language Tasks}
We used GPT-2 pre-training with SRD on the FineWeb-Edu dataset~\cite{penedo2024finewebdatasetsdecantingweb}. The results in the Table~\ref{tab:nlp_tasks_gpt2} show that SRD achieved substantial computational savings, reducing both the number of training steps and wall-clock time by more than half.
\begin{table}[h]
\centering
\caption{NLP task performance of SRD for small GPT-2 pretraining experiments.}
\label{tab:nlp_tasks_gpt2}
\begin{tabular}{l|c|c|c|c}
\toprule
\textbf{Model} & \textbf{PPL\_val} & \textbf{PPL\_train} & \textbf{Num\_steps} & \textbf{Wall clock} \\
\midrule
GPT-2 (Baseline) & $29.37$ & $28.50$ & $2,560,000$ & $13.34$ hrs \\
SRD (Threshold = 0.99) & $32.81$ & $31.78$ & $\mathbf{1,066,720}$ & $\mathbf{5.91}$ hrs \\
\bottomrule
\end{tabular}
\end{table}

\vspace{-0.3cm}
\subsection{Ablation Study}

To better understand the behavior of Progressive Data Dropout, we conduct an ablation study along three dimensions: dropout threshold, training duration, and matched-step baseline comparisons. For all our experiments, we use EfficientNet-B0 trained from scratch on CIFAR-100.

\textbf{Are our efficiency gains just due to fewer effective epochs?} To ensure our improvements are not simply due to reduced training time, we train a baseline EfficientNet-B0 for the same number of epochs as the equivalent effective epoch for each dropout variant. This lets us isolate whether performance improvements come from the dropout schedule itself rather than step count alone. We report these in Table~\ref{tab:matched_baseline}.

\textbf{How sensitive is difficulty-based dropout to the choice of threshold?} We vary the confidence threshold $\tau$ used for dropping samples in the difficulty-based variant (DBPD), ranging from 0.1 to 0.9 in increments of 0.1 and report these results in Table~\ref{tab:threshold_ablation} , Table~\ref{tab:cifar100_threshold_sensitivity}, and Table~\ref{tab:imagenet_threshold_sensitivity}. 
Higher thresholds lead to more conservative early-stage dropout, while lower values remove more data per epoch. Hence, lower thresholds tend to have a much lower step count. To ensure a balance between accuracy and epochs, we use thresholds of 0.3 and 0.7 in all experiments. Experiments in Table~\ref{tab:threshold_ablation} are conducted using EfficientNet-B0 on CIFAR-100 for 200 epochs with the final epoch being full-dataset revision. For SRD, we applied it in the most straightforward form of data dropping. Table~\ref{tab:srd_cifar100_sensitivity} presents the results on CIFAR-100 across different thresholds for reference.

%\textcolor{red}{Suggestion in the above paragraph, when our threshold is high more data points are retained in \textbf{early} stages. For instance, if the threshold is 0.7, when we compare it against individual outputs $\hat{y}$ in the early stages say epoch 1, there are a lot of chances that $\hat{y} < 0.7$, hence it gets retained for training. This can also be shown by how the step count for 0.7 has always been higher than 0.3, as well as the graphs that we have for samples dropped vs epochs. So, I think it should be "Higher thresholds lead to more conservative early-stage dropout, while the lower values liberate more data per epoch" }

\begin{table}[h]
\centering
\caption{Varying threshold on EfficientNet-B0 (CIFAR-100, 200 epochs). Threshold 0.0 = baseline.}
\resizebox{\linewidth}{!}{
\begin{tabular}{c|cccccccccc}
\toprule
\textbf{Threshold} & 0.0 & 0.1 & 0.2 & 0.3 & 0.4 & 0.5 & 0.6 & 0.7 & 0.8 & 0.9 \\
\midrule
\textbf{Accuracy (\%)} & 66.32 & 66.61 & 66.92 & 67.15 & 68.07 & 67.53 & 68.80 & 68.89 & \textbf{69.03} & 68.93 \\
\bottomrule
\end{tabular}
}
\label{tab:threshold_ablation}
\end{table}

\begin{table}[h]
\centering
\caption{Analysis for different dropout thresholds on CIFAR-100 dataset with training from scratch.}
\label{tab:cifar100_threshold_sensitivity}
\resizebox{\linewidth}{!}{
\begin{tabular}{l|c|c|c|c|c|c|c|c|c|c}
\toprule
\textbf{Threshold} & \textbf{0.0} & \textbf{0.1} & \textbf{0.2} & \textbf{0.3} & \textbf{0.4} & \textbf{0.5} & \textbf{0.6} & \textbf{0.7} & \textbf{0.8} & \textbf{0.9} \\
\midrule
EfficientFormer & $55.91$ & $56.37$ & $56.88$ & $57.52$ & $57.85$ & $57.98$ & $58.12$ & $58.08$ & $58.17$ & $58.22$ \\
MobileNet-v2 & $61.25$ & $61.49$ & $62.17$ & $62.85$ & $62.98$ & $63.21$ & $63.27$ & $63.31$ & $63.86$ & $63.79$ \\
\bottomrule
\end{tabular}}
\end{table}
\begin{table}[h]
\centering
\caption{Analysis for different dropout thresholds on ImageNet with training from scratch.}
\label{tab:imagenet_threshold_sensitivity}
\resizebox{\linewidth}{!}{
\begin{tabular}{l|c|c|c|c|c|c|c|c|c|c}
\toprule
\textbf{Threshold} & \textbf{0.0} & \textbf{0.1} & \textbf{0.2} & \textbf{0.3} & \textbf{0.4} & \textbf{0.5} & \textbf{0.6} & \textbf{0.7} & \textbf{0.8} & \textbf{0.9} \\
\midrule
ResNet-50 & $75.04$ & $75.55$ & $75.80$ & $76.21$ & $76.34$ & $76.42$ & $76.79$ & $77.14$ & $77.20$ & $77.19$ \\
\bottomrule
\end{tabular}}
\end{table}
\begin{table}[H]
\centering
\caption{SRD Threshold sensitivity on CIFAR-100 dataset.}
\label{tab:srd_cifar100_sensitivity}
\resizebox{\linewidth}{!}{
\begin{tabular}{l|c|c|c|c|c|c|c|c|c|c}
\toprule
\textbf{Threshold} & \textbf{0.0} & \textbf{0.2} & \textbf{0.4} & \textbf{0.6} & \textbf{0.8} & \textbf{0.9} & \textbf{0.92} & \textbf{0.94} & \textbf{0.95} & \textbf{0.96} \\
\midrule
EfficientFormer & $55.91$ & $17.84$ & $27.92$ & $38.85$ & $48.83$ & $56.15$ & $56.74$ & $56.91$ & $56.95$ & $56.97$ \\
MobileNet-v2 & $61.25$ & $24.41$ & $35.22$ & $48.63$ & $57.79$ & $62.18$ & $62.59$ & $62.77$ & $62.80$ & $62.82$ \\
\bottomrule
\end{tabular}}
\end{table}

\textbf{Can we shorten training and still retain performance?} To assess training-time sensitivity, we repeat all dropout variants under a 50-epoch training schedule (including final revision). We compare them to a 50-epoch baseline. Despite shorter training, all variants outperform the baseline (and the baseline trained to 200 epochs as well). Further, with just 50 epochs we are only 1.75\% worse off than our variant with 200 epochs. However, the baseline drops by 2.67\%. This shows that the proposed data dropout can improve generalization even under constrained budgets. The results can be seen in Table~\ref{tab:ablation_50epoch}.

\begin{table}[t]
\centering
\caption{Accuracy (\%) and effective epochs reported as \textit{accuracy/effective epochs} for CIFAR-100 with EfficientNet-B0, trained for 25 and 50 epochs from scratch.}
\resizebox{\linewidth}{!}{
\begin{tabular}{l|c|c|c|c|c|c}
\toprule
\textbf{Model } & \textbf{Baseline} & \textbf{DBPD 0.3} & \textbf{DBPD 0.7} & \textbf{SRD 0.95} & \textbf{SMRD 0.3} & \textbf{SMRD 0.7} \\
\midrule
(50 epochs) &&&&&& \\
EfficientNet-B0 (C100) & 63.65 / 50 & 67.65 / 14.48 & 67.62 / 18.64 & 68.32 / 24.8 & \textbf{69.26} / 16.76 & 67.44 / 21.8\\
\hline
(25 epochs)  &&&&&& \\
EfficientNet-B0 (C100) & 63.32 / 25 & \textbf{67.22} / \textbf{10.8} & 66.94 / 14.02 & 17.36 / 17.36 & 67.17 / 11.56 & 66.18 / 14.62\\
\bottomrule
\end{tabular}
}
\label{tab:ablation_50epoch}
\end{table}

\begin{figure}[htb]
    \centering
    \begin{minipage}{0.45\linewidth}
        \centering
        \includegraphics[width=\linewidth]{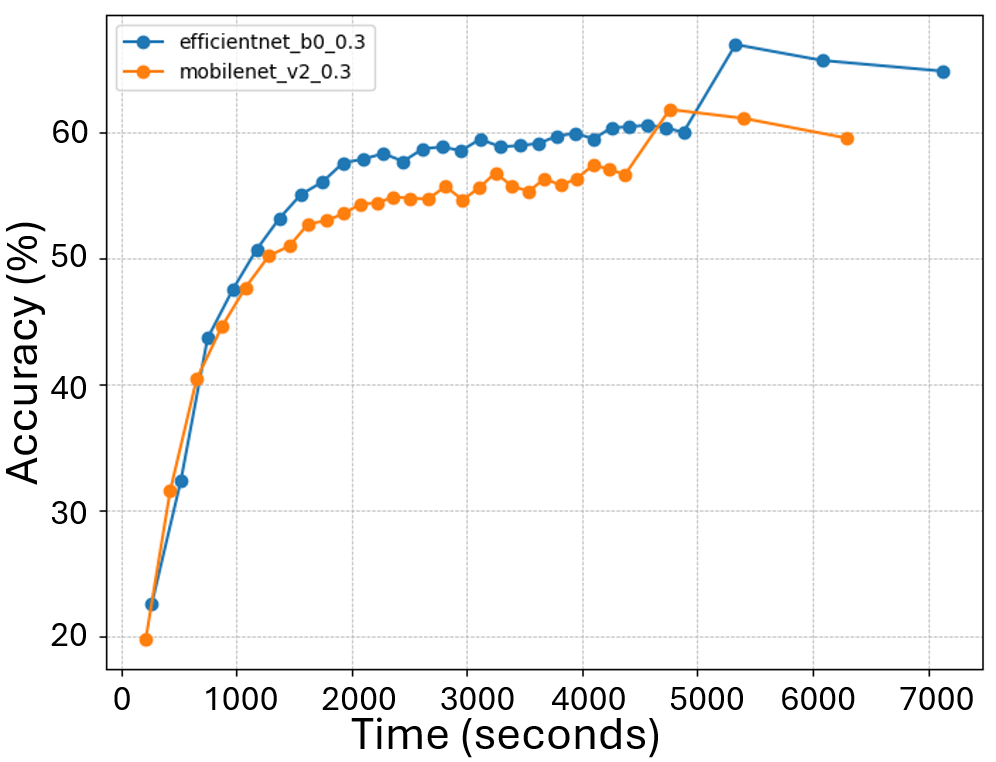}
        \caption{We evaluate EfficientNet and MobileNet by adding more than 1 epoch worth of revision. We see that after an initial boost, there is a drop in performance.}
        \label{fig:revision-moreepoch}
    \end{minipage}
    \hfill
    \begin{minipage}{0.45\linewidth}
        \centering
        \includegraphics[width=\linewidth]{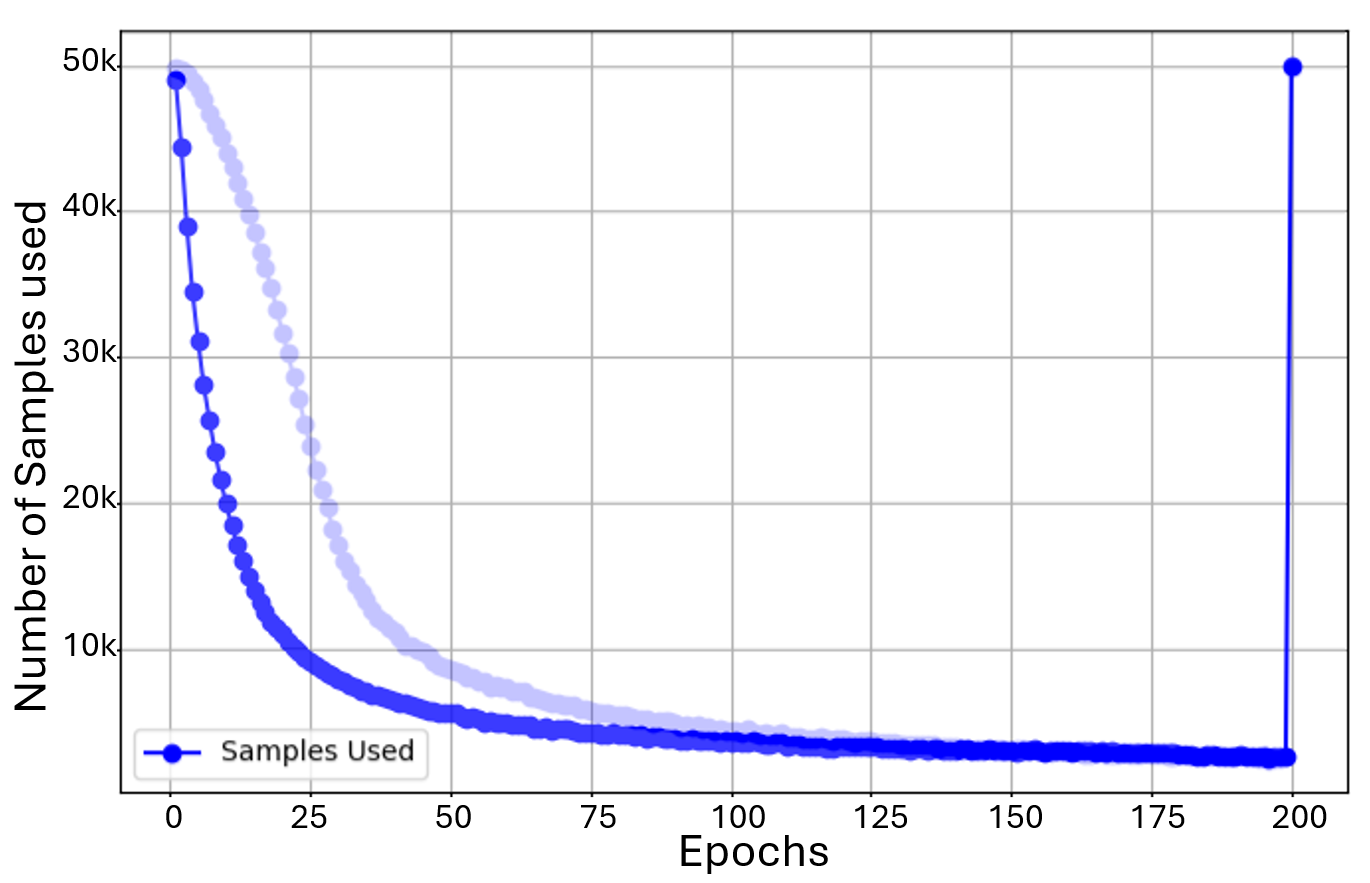}
        \caption{Here, we notice an exponential decay in the number of samples being picked at each epoch for backpropagation. The last epoch includes all samples again. The lighter plot denotes threshold 0.7 while the darker one represents 0.3.}
        \label{fig:dropped}
    \end{minipage}
\end{figure}

\textbf{What happens when we use more epochs for revision?} We observe that test performance often peaks shortly after the first revision epoch, followed by a gradual decline. As shown in Figure~\ref{fig:revision-moreepoch}, the model experiences a significant boost in accuracy during the first revision but this is subsequently followed by a steady drop, eventually converging toward the performance of the baseline model. We hypothesize that this pattern arises because the model, after the initial generalization gains, begins to overfit to the full dataset, thereby losing some of its earlier generalization capability.

\textbf{How many samples are we dropping over time?} 
Figure~\ref{fig:dropped} illustrates the number of training samples retained at each epoch under the DBPD (Difficulty-Based Progressive Dropout) strategy for two different threshold settings: 0.3 (darker line) and 0.7 (lighter line). In both cases, training begins with the full dataset. As the model improves and classifies more samples with high confidence, the number of retained examples drops sharply, particularly in the early epochs.
The higher threshold (0.7) causes a more aggressive pruning of data, dropping a larger fraction of confident samples per epoch. As a result, it reaches a low sample count much earlier, within the first 15–20 epochs, leaving fewer than 10,000 samples for much of training. In contrast, the 0.3 threshold maintains more samples for a longer duration, leading to a more gradual decay.
Both strategies conclude with a final epoch in which the full dataset is reintroduced for revision, producing a sharp spike back to 50,000 samples. This U-shaped trajectory reflects the design principle of DBPD: prioritize hard examples early, reduce redundant computation mid-training, and consolidate knowledge through a final full-dataset pass.

\section{Limitations}
\label{sec:limitations}

While Progressive Data Dropout demonstrates significant improvements in training efficiency and generalization, it has certain limitations. The method is most effective when training models from scratch; in fine-tuning or transfer learning scenarios, the relative gains are slightly lower. Whilst there are still some gains, pre-training helps to reduce that. Our current evaluation is limited to classification tasks, with preliminary extensions. However, broader applicability to other domains such as segmentation, language modeling, or multi-modal learning remains to be investigated. Additionally, we observe that larger datasets generally require more effective epochs and show lower gains, which may limit some of the training savings. Finally, the method introduces hyperparameters such as thresholds or decay rates, that can be sensitive to the dataset and model. While our default settings perform well across multiple benchmarks, optimal tuning may be necessary in more specialized contexts. Future work would be coming up with mathematical approximations to avoid the use of difficulty based scheduling and adapting the approach to different modalities and tasks.

\section{Conclusion}

We introduced \textit{Progressive Data Dropout}, a simple yet effective approach for improving training efficiency by progressively reducing the amount of data used in each epoch. Through three variants, difficulty-based, scalar-based, and schedule-matched random dropout, we demonstrated that substantial computational savings can be achieved without sacrificing, and in many cases even improving, generalization performance. Notably, the schedule-matched random variant consistently outperformed more structured approaches, highlighting the surprising strength of stochastic data reduction. Our method reduces effective epochs to as little as 12.4\% of standard training, with accuracy gains of up to 4.82\%. These results suggest that thoughtful data dropout scheduling can serve as a powerful tool for accelerating training in resource-constrained settings, without the need for architectural changes or complex model tuning. We hope this work inspires further exploration of data-centric regularization strategies.

\bibliography{neurips_2025}

\begin{thebibliography}{46}
\providecommand{\natexlab}[1]{#1}
\providecommand{\url}[1]{\texttt{#1}}
\expandafter\ifx\csname urlstyle\endcsname\relax
  \providecommand{\doi}[1]{doi: #1}\else
  \providecommand{\doi}{doi: \begingroup \urlstyle{rm}\Url}\fi

\bibitem[Balcan et~al.(2007)Balcan, Broder, and Zhang]{balcan2007margin}
Maria-Florina Balcan, Andrei Broder, and Tong Zhang.
\newblock Margin based active learning.
\newblock In \emph{International Conference on Computational Learning Theory}, pages 35--50. Springer, 2007.

\bibitem[Bengio et~al.(2009)Bengio, Louradour, Collobert, and Weston]{bengio2009curriculum}
Yoshua Bengio, J{\'e}r{\^o}me Louradour, Ronan Collobert, and Jason Weston.
\newblock Curriculum learning.
\newblock In \emph{Proceedings of the 26th annual international conference on machine learning}, pages 41--48, 2009.

\bibitem[Bjork et~al.(2011)Bjork, Bjork, et~al.]{bjork2011making}
Elizabeth~L Bjork, Robert~A Bjork, et~al.
\newblock Making things hard on yourself, but in a good way: Creating desirable difficulties to enhance learning.
\newblock \emph{Psychology and the real world: Essays illustrating fundamental contributions to society}, 2\penalty0 (59-68), 2011.

\bibitem[Cao et~al.(2019)Cao, Wei, Gaidon, Arechiga, and Ma]{cao2019learning}
Kaidi Cao, Colin Wei, Adrien Gaidon, Nikos Arechiga, and Tengyu Ma.
\newblock Learning imbalanced datasets with label-distribution-aware margin loss.
\newblock \emph{Advances in neural information processing systems}, 32, 2019.

\bibitem[Coleman et~al.()Coleman, Yeh, Mussmann, Mirzasoleiman, Bailis, Liang, Leskovec, and Zaharia]{colemanselection}
Cody Coleman, Christopher Yeh, Stephen Mussmann, Baharan Mirzasoleiman, Peter Bailis, Percy Liang, Jure Leskovec, and Matei Zaharia.
\newblock Selection via proxy: Efficient data selection for deep learning.
\newblock In \emph{International Conference on Learning Representations}.

\bibitem[Deng et~al.(2009)Deng, Dong, Socher, Li, Li, and Fei-Fei]{deng2009imagenet}
Jia Deng, Wei Dong, Richard Socher, Li-Jia Li, Kai Li, and Li~Fei-Fei.
\newblock Imagenet: A large-scale hierarchical image database.
\newblock In \emph{2009 IEEE Conference on Computer Vision and Pattern Recognition}, pages 248--255. IEEE, 2009.
\newblock \doi{10.1109/CVPR.2009.5206848}.

\bibitem[Dosovitskiy et~al.(2020)Dosovitskiy, Beyer, Kolesnikov, Weissenborn, Zhai, Unterthiner, Dehghani, Minderer, Heigold, Gelly, et~al.]{dosovitskiy2020image}
Alexey Dosovitskiy, Lucas Beyer, Alexander Kolesnikov, Dirk Weissenborn, Xiaohua Zhai, Thomas Unterthiner, Mostafa Dehghani, Matthias Minderer, Georg Heigold, Sylvain Gelly, et~al.
\newblock An image is worth 16x16 words: Transformers for image recognition at scale.
\newblock \emph{arXiv preprint arXiv:2010.11929}, 2020.

\bibitem[Dunlosky et~al.(2013)Dunlosky, Rawson, Marsh, Nathan, and Willingham]{dunlosky2013improving}
John Dunlosky, Katherine~A Rawson, Elizabeth~J Marsh, Mitchell~J Nathan, and Daniel~T Willingham.
\newblock Improving students’ learning with effective learning techniques: Promising directions from cognitive and educational psychology.
\newblock \emph{Psychological Science in the Public interest}, 14\penalty0 (1):\penalty0 4--58, 2013.

\bibitem[{Epoch AI}(2023)]{epoch2023aitrends}
{Epoch AI}.
\newblock Key trends and figures in machine learning, 2023.
\newblock URL \url{https://epoch.ai/trends}.
\newblock Accessed: 2025-10-19.

\bibitem[Feldman and Zhang(2020)]{feldman2020neural}
Vitaly Feldman and Chiyuan Zhang.
\newblock What neural networks memorize and why: Discovering the long tail via influence estimation.
\newblock \emph{Advances in Neural Information Processing Systems}, 33:\penalty0 2881--2891, 2020.

\bibitem[Gowda et~al.(2023)Gowda, Hao, Li, Gowda, Jin, and Sevilla-Lara]{gowda2023watt}
Shreyank~N Gowda, Xinyue Hao, Gen Li, Shashank~Narayana Gowda, Xiaobo Jin, and Laura Sevilla-Lara.
\newblock Watt for what: Rethinking deep learning's energy-performance relationship.
\newblock \emph{arXiv preprint arXiv:2310.06522}, 2023.

\bibitem[Han et~al.(2015)Han, Pool, Tran, and Dally]{han2015learning}
Song Han, Jeff Pool, John Tran, and William Dally.
\newblock Learning both weights and connections for efficient neural network.
\newblock \emph{Advances in neural information processing systems}, 28, 2015.

\bibitem[He et~al.(2016)He, Zhang, Ren, and Sun]{he2016deep}
Kaiming He, Xiangyu Zhang, Shaoqing Ren, and Jian Sun.
\newblock Deep residual learning for image recognition.
\newblock In \emph{Proceedings of the IEEE conference on computer vision and pattern recognition}, pages 770--778, 2016.

\bibitem[He et~al.(2022)He, Chen, Xie, Li, Doll{\'a}r, and Girshick]{he2022masked}
Kaiming He, Xinlei Chen, Saining Xie, Yanghao Li, Piotr Doll{\'a}r, and Ross Girshick.
\newblock Masked autoencoders are scalable vision learners.
\newblock In \emph{Proceedings of the IEEE/CVF conference on computer vision and pattern recognition}, pages 16000--16009, 2022.

\bibitem[Hinton et~al.(2015)Hinton, Vinyals, and Dean]{hinton2015distilling}
Geoffrey Hinton, Oriol Vinyals, and Jeff Dean.
\newblock Distilling the knowledge in a neural network.
\newblock \emph{arXiv preprint arXiv:1503.02531}, 2015.

\bibitem[Howard et~al.(2017)Howard, Zhu, Chen, Kalenichenko, Wang, Weyand, Andreetto, and Adam]{howard2017mobilenets}
Andrew~G Howard, Menglong Zhu, Bo~Chen, Dmitry Kalenichenko, Weijun Wang, Tobias Weyand, Marco Andreetto, and Hartwig Adam.
\newblock Mobilenets: Efficient convolutional neural networks for mobile vision applications.
\newblock \emph{arXiv preprint arXiv:1704.04861}, 2017.

\bibitem[Huang et~al.(2016)Huang, Sun, Liu, Sedra, and Weinberger]{huang2016deep}
Gao Huang, Yu~Sun, Zhuang Liu, Daniel Sedra, and Kilian~Q Weinberger.
\newblock Deep networks with stochastic depth.
\newblock In \emph{Computer Vision--ECCV 2016: 14th European Conference, Amsterdam, The Netherlands, October 11--14, 2016, Proceedings, Part IV 14}, pages 646--661. Springer, 2016.

\bibitem[Jacob et~al.(2018)Jacob, Kligys, Chen, Zhu, Tang, Howard, Adam, and Kalenichenko]{jacob2018quantization}
Benoit Jacob, Skirmantas Kligys, Bo~Chen, Menglong Zhu, Matthew Tang, Andrew Howard, Hartwig Adam, and Dmitry Kalenichenko.
\newblock Quantization and training of neural networks for efficient integer-arithmetic-only inference.
\newblock In \emph{Proceedings of the IEEE conference on computer vision and pattern recognition}, pages 2704--2713, 2018.

\bibitem[Katharopoulos and Fleuret(2018)]{katharopoulos2018not}
Angelos Katharopoulos and Fran{\c{c}}ois Fleuret.
\newblock Not all samples are created equal: Deep learning with importance sampling.
\newblock In \emph{International conference on machine learning}, pages 2525--2534. PMLR, 2018.

\bibitem[Kee et~al.(2018)Kee, Del~Castillo, and Runger]{kee2018query}
Seho Kee, Enrique Del~Castillo, and George Runger.
\newblock Query-by-committee improvement with diversity and density in batch active learning.
\newblock \emph{Information Sciences}, 454:\penalty0 401--418, 2018.

\bibitem[Koh and Liang(2017)]{koh2017understanding}
Pang~Wei Koh and Percy Liang.
\newblock Understanding black-box predictions via influence functions.
\newblock In \emph{International conference on machine learning}, pages 1885--1894. PMLR, 2017.

\bibitem[Krizhevsky and Hinton(2009)]{krizhevsky2009learning}
Alex Krizhevsky and Geoffrey Hinton.
\newblock Learning multiple layers of features from tiny images.
\newblock Technical report, University of Toronto, 2009.
\newblock URL \url{https://www.cs.toronto.edu/~kriz/learning-features-2009-TR.pdf}.

\bibitem[Li et~al.(2022)Li, Yuan, Wen, Hu, Evangelidis, Tulyakov, Wang, and Ren]{li2022efficientformer}
Yanyu Li, Geng Yuan, Yang Wen, Ju~Hu, Georgios Evangelidis, Sergey Tulyakov, Yanzhi Wang, and Jian Ren.
\newblock Efficientformer: Vision transformers at mobilenet speed.
\newblock \emph{Advances in Neural Information Processing Systems}, 35:\penalty0 12934--12949, 2022.

\bibitem[Lin et~al.(2017)Lin, Goyal, Girshick, He, and Doll{\'a}r]{lin2017focal}
Tsung-Yi Lin, Priya Goyal, Ross Girshick, Kaiming He, and Piotr Doll{\'a}r.
\newblock Focal loss for dense object detection.
\newblock In \emph{Proceedings of the IEEE international conference on computer vision}, pages 2980--2988, 2017.

\bibitem[Mindermann et~al.(2022)Mindermann, Brauner, Razzak, Sharma, Kirsch, Xu, H{\"o}ltgen, Gomez, Morisot, Farquhar, et~al.]{mindermann2022prioritized}
S{\"o}ren Mindermann, Jan~M Brauner, Muhammed~T Razzak, Mrinank Sharma, Andreas Kirsch, Winnie Xu, Benedikt H{\"o}ltgen, Aidan~N Gomez, Adrien Morisot, Sebastian Farquhar, et~al.
\newblock Prioritized training on points that are learnable, worth learning, and not yet learnt.
\newblock In \emph{International Conference on Machine Learning}, pages 15630--15649. PMLR, 2022.

\bibitem[Mirzasoleiman et~al.(2020)Mirzasoleiman, Bilmes, and Leskovec]{mirzasoleiman2020coresets}
Baharan Mirzasoleiman, Jeff Bilmes, and Jure Leskovec.
\newblock Coresets for data-efficient training of machine learning models.
\newblock In \emph{International Conference on Machine Learning}, pages 6950--6960. PMLR, 2020.

\bibitem[Nguyen et~al.(2022)Nguyen, Shaker, and H{\"u}llermeier]{nguyen2022measure}
Vu-Linh Nguyen, Mohammad~Hossein Shaker, and Eyke H{\"u}llermeier.
\newblock How to measure uncertainty in uncertainty sampling for active learning.
\newblock \emph{Machine Learning}, 111\penalty0 (1):\penalty0 89--122, 2022.

\bibitem[Paul et~al.(2021)Paul, Ganguli, and Dziugaite]{paul2021deep}
Mansheej Paul, Surya Ganguli, and Gintare~Karolina Dziugaite.
\newblock Deep learning on a data diet: Finding important examples early in training.
\newblock \emph{Advances in neural information processing systems}, 34:\penalty0 20596--20607, 2021.

\bibitem[Penedo et~al.(2024)Penedo, Kydlíček, allal, Lozhkov, Mitchell, Raffel, Werra, and Wolf]{penedo2024finewebdatasetsdecantingweb}
Guilherme Penedo, Hynek Kydlíček, Loubna~Ben allal, Anton Lozhkov, Margaret Mitchell, Colin Raffel, Leandro~Von Werra, and Thomas Wolf.
\newblock The fineweb datasets: Decanting the web for the finest text data at scale, 2024.
\newblock URL \url{https://arxiv.org/abs/2406.17557}.

\bibitem[Qin et~al.()Qin, Wang, Zheng, Gu, Peng, Zhou, Shang, Sun, Xie, You, et~al.]{qininfobatch}
Ziheng Qin, Kai Wang, Zangwei Zheng, Jianyang Gu, Xiangyu Peng, Daquan Zhou, Lei Shang, Baigui Sun, Xuansong Xie, Yang You, et~al.
\newblock Infobatch: Lossless training speed up by unbiased dynamic data pruning.
\newblock In \emph{The Twelfth International Conference on Learning Representations}.

\bibitem[Rahmani and Atia(2018)]{rahmani2018data}
Mostafa Rahmani and George~K Atia.
\newblock Data dropout in arbitrary basis for deep network regularization.
\newblock In \emph{2018 52nd Asilomar Conference on Signals, Systems, and Computers}, pages 66--70. IEEE, 2018.

\bibitem[Sandler et~al.(2018)Sandler, Howard, Zhu, Zhmoginov, and Chen]{sandler2018mobilenetv2}
Mark Sandler, Andrew Howard, Menglong Zhu, Andrey Zhmoginov, and Liang-Chieh Chen.
\newblock Mobilenetv2: Inverted residuals and linear bottlenecks.
\newblock In \emph{Proceedings of the IEEE conference on computer vision and pattern recognition}, pages 4510--4520, 2018.

\bibitem[Settles(2009)]{settles2009active}
Burr Settles.
\newblock Active learning literature survey.
\newblock 2009.

\bibitem[Shrivastava et~al.(2016)Shrivastava, Gupta, and Girshick]{shrivastava2016training}
Abhinav Shrivastava, Abhinav Gupta, and Ross Girshick.
\newblock Training region-based object detectors with online hard example mining.
\newblock In \emph{Proceedings of the IEEE conference on computer vision and pattern recognition}, pages 761--769, 2016.

\bibitem[Srivastava et~al.(2014)Srivastava, Hinton, Krizhevsky, Sutskever, and Salakhutdinov]{srivastava2014dropout}
Nitish Srivastava, Geoffrey Hinton, Alex Krizhevsky, Ilya Sutskever, and Ruslan Salakhutdinov.
\newblock Dropout: a simple way to prevent neural networks from overfitting.
\newblock \emph{The journal of machine learning research}, 15\penalty0 (1):\penalty0 1929--1958, 2014.

\bibitem[Tan and Le(2019)]{tan2019efficientnet}
Mingxing Tan and Quoc Le.
\newblock Efficientnet: Rethinking model scaling for convolutional neural networks.
\newblock In \emph{International conference on machine learning}, pages 6105--6114. PMLR, 2019.

\bibitem[Toneva et~al.()Toneva, Sordoni, des Combes, Trischler, Bengio, and Gordon]{tonevaempirical}
Mariya Toneva, Alessandro Sordoni, Remi~Tachet des Combes, Adam Trischler, Yoshua Bengio, and Geoffrey~J Gordon.
\newblock An empirical study of example forgetting during deep neural network learning.
\newblock In \emph{International Conference on Learning Representations}.

\bibitem[Wan et~al.(2013)Wan, Zeiler, Zhang, Le~Cun, and Fergus]{wan2013regularization}
Li~Wan, Matthew Zeiler, Sixin Zhang, Yann Le~Cun, and Rob Fergus.
\newblock Regularization of neural networks using dropconnect.
\newblock In \emph{International conference on machine learning}, pages 1058--1066. PMLR, 2013.

\bibitem[Wang et~al.(2018)Wang, Huan, and Li]{wang2018data}
Tianyang Wang, Jun Huan, and Bo~Li.
\newblock Data dropout: Optimizing training data for convolutional neural networks.
\newblock In \emph{2018 IEEE 30th international conference on tools with artificial intelligence (ICTAI)}, pages 39--46. IEEE, 2018.

\bibitem[Wei et~al.(2021)Wei, Zhu, Cheng, Liu, Niu, and Liu]{wei2021learning}
Jiaheng Wei, Zhaowei Zhu, Hao Cheng, Tongliang Liu, Gang Niu, and Yang Liu.
\newblock Learning with noisy labels revisited: A study using real-world human annotations.
\newblock \emph{arXiv preprint arXiv:2110.12088}, 2021.

\bibitem[Wightman(2019)]{timm}
Ross Wightman.
\newblock Pytorch image models.
\newblock \url{https://github.com/rwightman/pytorch-image-models}, 2019.

\bibitem[Yang et~al.(2020)Yang, Tang, Yue, Yang, and Xu]{yang2020accelerating}
Naisen Yang, Hong Tang, Jianwei Yue, Xin Yang, and Zhihua Xu.
\newblock Accelerating the training process of convolutional neural networks for image classification by dropping training samples out.
\newblock \emph{IEEE Access}, 8:\penalty0 142393--142403, 2020.

\bibitem[Yang et~al.(2022)Yang, Xie, Peng, Xu, Sun, and Li]{yang2022dataset}
Shuo Yang, Zeke Xie, Hanyu Peng, Min Xu, Mingming Sun, and Ping Li.
\newblock Dataset pruning: Reducing training data by examining generalization influence.
\newblock \emph{arXiv preprint arXiv:2205.09329}, 2022.

\bibitem[Yang et~al.(2016)Yang, Jin, Tao, Xie, and Feng]{yang2016dropsample}
Weixin Yang, Lianwen Jin, Dacheng Tao, Zecheng Xie, and Ziyong Feng.
\newblock Dropsample: A new training method to enhance deep convolutional neural networks for large-scale unconstrained handwritten chinese character recognition.
\newblock \emph{Pattern Recognition}, 58:\penalty0 190--203, 2016.

\bibitem[Yuan et~al.(2025)Yuan, Lin, Feng, Han, and Liu]{yuan2025instancedependent}
Suqin Yuan, Runqi Lin, Lei Feng, Bo~Han, and Tongliang Liu.
\newblock Instance-dependent early stopping.
\newblock In \emph{The Thirteenth International Conference on Learning Representations}, 2025.

\bibitem[Zhong et~al.(2021)Zhong, Deng, Fang, Hu, Zhao, Li, and Wen]{zhong2021dynamic}
Yaoyao Zhong, Weihong Deng, Han Fang, Jiani Hu, Dongyue Zhao, Xian Li, and Dongchao Wen.
\newblock Dynamic training data dropout for robust deep face recognition.
\newblock \emph{IEEE Transactions on Multimedia}, 24:\penalty0 1186--1197, 2021.

\end{thebibliography}
\bibliographystyle{plainnat}

%%%%%%%%%%%%%%%%%%%%%%%%%%%%%%%%%%%%%%%%%%%%%%%%%%%%%%%%%%%%
\section*{Broader Impact}

This work proposes a simple and general-purpose strategy for improving the training efficiency of deep neural networks through selective data dropout. The proposed methods, difficulty-based, random, and schedule-matched dropout, reduce the number of training examples that undergo backpropagation while maintaining or improving final model accuracy. This can lead to significant reductions in computational cost and energy consumption during training.

\textbf{Positive societal impact.} By reducing training cost without modifying model architectures or optimizers, our approach lowers the barrier to entry for researchers and practitioners with limited access to high-performance computing resources. This has the potential to democratize access to deep learning research and make machine learning more environmentally sustainable. Furthermore, the method is easy to integrate with existing training pipelines and is applicable to a wide range of model families and datasets, promoting broader adoption of energy-efficient training practices.

\textbf{Potential risks.} Our experiments are currently focused on image classification tasks using benchmark datasets. In more sensitive or imbalanced domains (e.g., healthcare or fairness-critical applications), selectively dropping data, even temporarily, could unintentionally bias model learning or overlook rare but important examples. Although our dropout methods reintroduce the full dataset at the end of training, further work is needed to ensure that performance is equitable across subgroups in real-world settings. Additionally, while our method offers efficiency gains, it does not guarantee robustness against adversarial data distributions.

\textbf{Environmental impact.} The method is designed to reduce the number of effective epochs required to train a model to convergence. We report gains of up to 10x reduction in backpropagation steps, which translate to real-world energy savings, especially in large-scale training scenarios. This aligns with growing interest in making machine learning more environmentally responsible.

In summary, our work offers a promising and practical step toward efficient and accessible deep learning. We encourage future work to investigate its impact on underrepresented data groups and its integration with fairness-aware or privacy-preserving training pipelines.

\appendix

\section{Results for 0.7 variants}

In the main paper, we report results on the 0.3 variants since that is our best performing variant in most cases along with being more efficient in terms of effective epochs. For completeness, we also report results for the 0.7 threshold in Table~\ref{tab:cifar-0.7}. Again, we report results as Accuracy/Effective Epochs.

\begin{table}[htb]
\centering
\caption{Accuracy (\%) and effective epochs for threshold 0.7 variants. * refers to the model being pre-trained on ImageNet, else the model is trained from scratch.}
\begin{tabular}{l|c|c}
\toprule
\textbf{Model} & \textbf{DBPD 0.7} & \textbf{SMRD 0.7} \\
\midrule
EfficientNet-B0 (C10) & 90.01 / 9.16 & 89.87 / 9.72 \\
EfficientNet-B0 (C100) & 68.80 / 33 & 70.75 / 46.4 \\
*EfficientNet-B0 (C100) & 84.42 / 41 & 84.67 / 51 \\
%EfficientNet-B0 (I) & 78.89 / 200.2M & 79.15 / 259.5M \\
\midrule
MobileNet-V2 (C10) & 87.44 / 9.08 & 87.62 / 10.42 \\
MobileNet-V2 (C100) & 63.31 / 40.8 & 65.73 / 55.2 \\
*MobileNet-V2 (C100) & 74.58 / 40.6 & 75.23 / 47.2 \\
%MobileNet-V2 (I) & 70.26 / 98.42M & 72.77 / 164.5M \\
\midrule
ResNet-50 (C10) & 88.20 / 7.12 & 87.01 / 9.8 \\
ResNet-50 (C100) & 60.22 / 21 & 62.62 / 37.6 \\
*ResNet-50 (C100) & 80.51 / 35.6 & 81.45 / 42.8 \\
%ResNet-50 (I) & 77.05 / 67.5M & 79.11 / 75.5M \\
\midrule
EfficientFormer-L1 (C10) & 80.66 / 15.52 & 80.81 / 15.62 \\
EfficientFormer-L1 (C100) & 58.08 / 39 & 59.85 / 62.4 \\
*EfficientFormer-L1 (C100) & 86.89 / 59 & 87.78 / 76.8 \\
%EfficientFormer-L1 (I) & 79.45 / 141.5M & 80.08 / 159.6M \\
\midrule
*ViT-B-MAE (C100) & 86.94 / 71.8 & 87.45 / 82.2 \\
%ViT-B-MAE (I) & 83.18 / 58.5M & 84.01 / 66.3M \\
\bottomrule
\end{tabular}
\label{tab:cifar-0.7}
\end{table}

\section{Algorithm}

We provide a unified training algorithm (Algorithm~\ref{alg:revision}) that captures all three variants of Progressive Data Dropout: Difficulty-Based Progressive Dropout (DBPD), Scalar Random Dropout (SRD), and Schedule-Matched Random Dropout (SMRD). The shared logic between DBPD and SMRD is highlighted in \textcolor{blue}{blue}, the SRD-specific logic in \textcolor{violet}{violet}, and SMRD’s additional stochasticity in \textcolor{red}{red}.

\begin{algorithm}[htb]
\SetAlgoLined
\DontPrintSemicolon
\SetAlFnt{\small}
\SetAlCapFnt{\small}
\SetAlCapNameFnt{\small}
\SetAlgoSkip{bigskip}
\caption{Unified algorithm for all 3 variants, the common steps between DBPD and SMRD are in \textcolor{blue}{blue}, specific steps for SRD is in \textcolor{violet}{violet} and steps for SMRD is in \textcolor{red}{red}}
\KwIn{
Model $f_\theta$ with parameters $\theta$, Training data $\mathcal{D}_{\text{train}}$, test data $\mathcal{D}_{\text{test}}$, Number of epochs $E$, revision start epoch $E_{\text{rev}}$, \textcolor{blue}{Confidence threshold $\tau : 0 \leq \tau \leq 1$}, Loss function $\mathcal{L}$\textcolor{violet}{Decay factor : $0 < \gamma < 1$}
}
% \KwOut{Trained model $f_\theta$}

\For{epoch $= 1$ \KwTo $E$}{
    \eIf{epoch $\leq E_{\text{rev}}$}{
        \If{SRD}{\textcolor{violet}{$r \leftarrow \gamma^{epoch - 1}$}}
        ~\\
        \ForEach{batch $(\mathbf{X}, \mathbf{y}) \in \mathcal{D}_{\text{train}}$}{
            \If{SRD}{\textcolor{violet}{$n \leftarrow $ number of samples in $\textbf{X}$}\;
            \textcolor{violet}{$k \leftarrow [r \cdot n]$}\;
            \textcolor{violet}{\If{$k = 0$}{continue\;}}
            \textcolor{violet}{Randomly sample $k$ indices: $\mathcal{M} \subseteq {1,...,n}$}\;}
            \If{DBPD or SMRD}{
            \textcolor{blue}{$\hat{\mathbf{y}} = f_\theta(\mathbf{X})$\;
            \eIf{$\tau = 0$}{ 
                $\mathcal{M} \leftarrow \{i \mid \arg\max \hat{y}_i \neq y_i\}$\;
            }{
                $p_i = \max \sigma(\hat{\mathbf{y}}_i)$\;
                $\mathcal{M} \leftarrow \{i \mid p_i < \tau\}$\;
            }}}
            \If{$\mathcal{M} \neq \emptyset$}{
                \If{SMRD}{
                \textcolor{red}{Randomly sample $|\mathcal{M}|$ indices from ${0,...,|\mathbf{X}|-1}$}\;}
                $\theta \leftarrow optimizer(\eta, \theta, \mathcal{L}(f_\theta(\mathbf{X}_\mathcal{M}), \mathbf{y}_\mathcal{M})$\;
            }
        }
    }{
        \ForEach{batch $(\mathbf{X}, \mathbf{y}) \in \mathcal{D}_{\text{train}}$}{
            Standard training: 
            $\theta \leftarrow \theta - \eta \nabla_\theta \mathcal{L}(f_\theta(\mathbf{X}), \mathbf{y})$\;
        }
    }
}
\label{alg:revision}
\end{algorithm}
% \fi
\section{Mathematical Approximation for SMRD}
\label{sec:math_approx_smrd}
Among the three variants of Progressive Data Dropout we explore, the Schedule-Matched Random Dropout (SMRD) variant achieves the strongest performance across datasets and architectures. SMRD matches the per-epoch data retention schedule of the difficulty-based method (DBPD), but applies dropout stochastically rather than using sample confidence. This yields both generalization benefits and practical simplicity, but also introduces a dependency on an external schedule derived from a separate DBPD run.

To eliminate this dependency and make SMRD easier to use and evaluate independently, we propose a family of simple mathematical functions that approximate the number of training samples used at each epoch. These approximations allow us to:
\begin{itemize}
    \item Generate a complete dropout schedule without running DBPD.
    \item Integrate SMRD seamlessly into standard data loaders.
    \item Enable reproducible and lightweight evaluation of training efficiency.
\end{itemize}

We approximate the SMRD schedule using a function $f(x; \alpha)$, where $x$ is the epoch number and $\alpha$ controls the decay rate. To maintain full-dataset training in the final epoch, we define:

\[
f(x) = 
\begin{cases}
\hat{f}(x; \alpha), & \text{if } x < T \\
50{,}000, & \text{if } x = T
\end{cases}
\]

Here, $T$ is the total number of training epochs (e.g., $T = 200$), and $\hat{f}(x; \alpha)$ is one of the following parameterized decay functions, each normalized to output a maximum of $50{,}000$ (Using CIFAR-100 as our example) samples:

\begin{align*}
\text{(1) Power-Law Decay:} \quad & \hat{f}(x; \alpha) = \frac{1}{x^\alpha} \\
\text{(2) Exponential Decay:} \quad & \hat{f}(x; \alpha) = e^{-\alpha x} \\
\text{(3) Logarithmic Decay:} \quad & \hat{f}(x; \alpha) = \frac{1}{\log(x + \alpha)} \\
\text{(4) Inverse Linear Decay:} \quad & \hat{f}(x; \alpha) = \frac{1}{x + \alpha} \\
\text{(5) Sigmoid-Complement Decay:} \quad & \hat{f}(x; \alpha) = 1 - \frac{1}{1 + e^{-\alpha x}}
\end{align*}

Each function captures a different decay behavior, and the choice of $\alpha$ allows for close approximation of SMRD thresholds (e.g., $0.3$ and $0.7$) using a single parameter. Lower $\alpha$ values lead to gentler decay (slower dropout), while higher $\alpha$ values correspond to more aggressive early-stage data reduction.

These approximations not only reduce the reliance on a difficulty-based schedule but also make SMRD deployment fully self-contained and differentiable, facilitating future theoretical analysis and integration into automated training pipelines.

\section{Evaluating the approximations on CIFAR-10 and CIFAR-100}

We evaluate these approximations on CIFAR-100 and train EfficientNet-B0 from scratch for 50 epochs. Barring the Inverse Linear Function, we notice improvements over the baseline in all cases. However, SMRD 0.3 still gives us our best results. We do note the performance of Logarithmic Decay and consider more detailed ablations on this for future work. The results can be seen in Table~\ref{tab:diff_func_50epoch}.

\begin{table}[t]
\centering
\caption{Accuracy (\%) and effective epochs reported as \textit{accuracy/effective epochs} for CIFAR-100 with EfficientNet-B0, 50 epochs from scratch on various scheduler functions.}
\begin{tabular}{l|c|c}
\toprule
\textbf{Model } & \textbf{Accuracy} & \textbf{Effective Epochs} \\
\midrule
Baseline & 63.65 & 50\\
Inverse Linear Function & 63.51 & \textbf{9.4} \\
Logarithmic Decay & 68.62 & 19.12 \\
SMRD 0.3 & \textbf{69.26} & 16.76 \\
SMRD 0.7 & 67.44 & 21.8 \\
SRD 0.95 & 68.32 & 24.8 \\
\bottomrule
\end{tabular}

\label{tab:diff_func_50epoch}
\end{table}

\textbf{Do we still have consistent improvements if we are having a longer schedule?} We conducted additional experiments using longer training runs (200 epochs) on CIFAR-10 with EfficientNet-B0, ResNet-50, and MobileNet-v2 architectures. To ensure fairness, we used uniform optimization parameters across all models. Table~\ref{tab:imagenet_finetune_200} show that our method achieves consistent improvements under the longer training schedule.

% \begin{table}
% \centering
% \begin{minipage}{0.49\linewidth}
% \input{Tables_from_rebuttal/cifar10_200epochs}
% \label{tab:cifar_200epochs_mini}
% \end{minipage}
% \hfill
% \begin{minipage}{0.49\linewidth}
% \input{Tables_from_rebuttal/finetune_imagenet_200epochs}
% \label{tab:imagenet_200epochs_mini}
% \end{minipage}
% \end{table}

\begin{table}
\centering
\caption{Fine-tuning models pretrained on ImageNet (200 epochs).}
\label{tab:imagenet_finetune_200}
% \resizebox{\linewidth}{!}{
\begin{tabular}{l|c|c|c|c}
\toprule
\textbf{Model} & \textbf{Baseline} & \textbf{DBPD (0.3)} & \textbf{SRD (0.99)} & \textbf{SMRD (0.3)} \\
\midrule
EfficientNet-B0 &  95.55 / 200  &  96.11 / 25  &  96.40 / 84  &  \textbf{96.83 / 31}  \\
ResNet-50 &  96.41 / 200  &  96.73 / 27  &  96.83 / 84  &  \textbf{97.71 / 33}  \\
MobileNet-v2 &  93.91 / 200  &  94.41 / 26  &  95.12 / 84  &  \textbf{95.55 / 33}  \\
\bottomrule
\end{tabular}
% }
\end{table}

\textbf{What happens if we use strong regularization or augmentation method?} Using stronger regularization or augmentation reduces the impact of the current threshold settings in our method. As mentioned in the paper, we report results with a threshold of 0.3 to achieve a balance between speed and accuracy. However, if accuracy is prioritized, higher thresholds can be used, as demonstrated in the table. These results are based on CIFAR-10 training for 200 epochs, using CutMix style augmentation with the parameter $\alpha$ set to 1.
\begin{table}
\centering
\caption{Performance under CutMix augmentation.}
\label{tab:cutmix_performance_cifar10}
\resizebox{\linewidth}{!}{
\begin{tabular}{l|c|c|c|c|c|c}
\toprule
\textbf{Model} & \textbf{Baseline} & \textbf{DBPD(0.3)} & \textbf{DBPD(0.8)} & \textbf{SRD(0.99)} & \textbf{SMRD(0.3)} & \textbf{SMRD(0.8)} \\
\midrule
MobileNet v2 & $90.41/200$ & $90.96/22$ & $91.45/32$ & $\mathbf{93.29/84}$ & $91.41/26$ & $93.24/42$ \\
\bottomrule
\end{tabular}}
\end{table}

\textbf{How does increasing revision epochs or extending total epochs affect?} As shown in Figure~\ref{fig:revision-moreepoch}, adding more revision epochs causes overfitting and can degrade test accuracy. To validate this further, we tested SMRD with 25, 50, and 75 revision epochs (compared to the standard setting of 1). While these settings brought the accuracy closer to the baseline, they did not yield any improvements and risked overfitting as shown in Table~\ref{tab:smrd_revision_epochs}. Additionally, we increased the overall number of training epochs to bring the overall computation closer to the baseline. As shown in Table~\ref{tab:extending_training_epochs}, this did not really improve our performance much.

\begin{table}
\centering
\begin{minipage}{0.49\linewidth}
% \begin{table}
\centering
\caption{Results of SMRD by increasing revision epochs.}
% \label{tab:smrd_revision_epochs}
\resizebox{\linewidth}{!}{
\begin{tabular}{l|c|c}
\toprule
\textbf{Method} & \textbf{Revision Epochs} & \textbf{Accuracy (\%)} \\
\midrule
SMRD (standard) & $1$ & $71.01$ \\
SMRD & $25$ & $68.55$ \\
SMRD & $50$ & $66.35$ \\
SMRD & $75$ & $66.27$ \\
Baseline & --- & $66.32$ \\
\bottomrule
\end{tabular}}
% \end{table}
\label{tab:smrd_revision_epochs}
\end{minipage}
\hfill
\begin{minipage}{0.49\linewidth}
% \begin{table}
\centering
\caption{Results of extending total training epochs.}
% \label{tab:extending_training_epochs}
\resizebox{\linewidth}{!}{
\begin{tabular}{l|c|c|c}
\toprule
\textbf{Method} & \textbf{Total Epochs} & \textbf{Accuracy (\%)} & \textbf{Effective Epochs (EE)} \\
\midrule
DBPD & $200$ & $\mathbf{67.15}$ & $25$ \\
DBPD & $300$ & $67.09$ & $29$ \\
DBPD & $400$ & $66.28$ & $33$ \\
DBPD & $500$ & $66.59$ & $36$ \\
SMRD & $200$ & $\mathbf{71.01}$ & $35$ \\
SMRD & $300$ & $70.82$ & $44$ \\
SMRD & $400$ & $70.75$ & $49$ \\
SMRD & $500$ & $70.99$ & $51$ \\
Baseline & $200$ & $66.32$ & $200$ \\
Baseline & $300$ & $\mathbf{66.45}$ & $200$ \\
Baseline & $400$ & $66.37$ & $200$ \\
Baseline & $500$ & $66.35$ & $200$ \\
\bottomrule
\end{tabular}}
% \end{table}
\label{tab:extending_training_epochs}
\end{minipage}
\end{table}

\textbf{What happens if the dataset has class imbalance or noisy labels?} It is important to emphasize that PDD fundamentally focuses on computational efficiency rather than robustness to class imbalance or noise. Table~\ref{tab:long_tail_cifar10} presents results for Long-Tail Classification on CIFAR-10 with $\rho$ = 100, where $\rho$ denotes the ratio between the sample sizes of the most frequent and least frequent classes. The dataset is generated following the same procedure as LDAM-DRW~\cite{cao2019learning}. We use Focal Loss~\cite{lin2017focal}, consistent with the baseline, to ensure a fair comparison. From the table, all PDD variants show substantial gains over the baseline, with SMRD (0.3) emerging as the best-performing method under class imbalance, achieving the highest accuracy. However, compared to the standard setting, this configuration requires more EE. We evaluated our method on noisy label classification using the Noisy CIFAR-10 dataset, following the setup from~\cite{wei2021learning}, with a noise rate of 17.3\% and random noise type, training for 50 epochs. Table~\ref{tab:label_noise_cifar10} shows that all three PDD variants outperform the baseline in both accuracy and efficiency. Notably, SRD (0.95) achieves the highest accuracy, though with slightly higher EE. SMRD and DBPD provide a better trade-off, offering much lower EE while maintaining strong performance, demonstrating their robustness under label noise.

\begin{table}
\centering
\begin{minipage}{0.49\linewidth}
% \begin{table}
\centering
\caption{Long-tail classification of ResNet-34 on CIFAR-10.}
% \label{tab:long_tail_cifar10}
\resizebox{\linewidth}{!}{
\begin{tabular}{l|c|c}
\toprule
\textbf{Model} & \textbf{Accuracy (\%)} & \textbf{Effective Epochs (EE)} \\
\midrule
Baseline &  46.04 & 200  \\
DBPD (0.3) &  58.09 & 64  \\
DBPD (0.2) &  54.82 & 55  \\
DBPD (0.1) &  24.28 & 65  \\
SRD (0.95) &  54.28 & 84  \\
SMRD (0.3) &  \textbf{60.14} &  \textbf{58}  \\
\bottomrule
\end{tabular}}
% \end{table}
\label{tab:long_tail_cifar10}
\end{minipage}
\hfill
\begin{minipage}{0.49\linewidth}
% \begin{table}[h]
\centering
\caption{Results of noisy label classification on Noisy CIFAR-10}
% \label{tab:label_noise_cifar10}
\resizebox{\linewidth}{!}{
\begin{tabular}{l|c|c}
\toprule
\textbf{Model} & \textbf{Accuracy (\%)} & \textbf{Effective Epochs (EE)} \\
\midrule
Baseline &  71.91  &  50  \\
DBPD (0.3) &  74.26  &  9  \\
SRD (0.95) &  \textbf{75.23}  &  \textbf{18}  \\
SMRD (0.3) &  73.76  &  9  \\
\bottomrule
\end{tabular}}
% \end{table}
\label{tab:label_noise_cifar10}
\end{minipage}
\end{table}

\textbf{What is the ideal dataset characteristics for PDD?} PDD is particularly beneficial for mid-sized to large datasets, where the computational savings are most significant. We also evaluated it on smaller fine-grained datasets, such as CUB and Flowers, and observed consistent computational savings without any loss in accuracy as shown in Table~\ref{tab:small_dataset_performance}. These results highlight PDD’s suitability across different dataset scales and complexities.

\begin{table}[h]
\centering
\caption{Model performance comparison on smaller fine-grained datasets.}
\label{tab:small_dataset_performance}
\begin{tabular}{l|l|c|c}
\toprule
\textbf{Model} & \textbf{Dataset} & \textbf{SMRD (0.3)} & \textbf{Baseline} \\
\midrule
EfficientNet-B0 & Airplane &  \textbf{80.07 / 9}  &  78.06 / 100  \\
EfficientNet-B0 & CUB &  \textbf{71.03 / 9}  &  69.83 / 100  \\
MobileNet-v2 & Airplane &  \textbf{76.92 / 10}  &  73.87 / 100  \\
MobileNet-v2 & CUB &  \textbf{68.94 / 10}  &  67.45 / 100  \\
\bottomrule
\end{tabular}
\end{table}

\textbf{How do the three variants perform given the same fixed compute budget?} We conducted a preliminary experiment on CIFAR-100 using ResNet-50, aligning the EE across all methods. All models were fine-tuned from ImageNet pre-training. Table~\ref{tab:aligning_ee_cifar100} shows that even under equal EE, all PDD variants outperform the baseline, with SMRD remaining the best-performing variant.

\begin{table}
\centering
\caption{Performance of aligning Effective Epochs (EE) across methods on CIFAR-100 (ResNet-50)}
\label{tab:aligning_ee_cifar100}
\begin{tabular}{l|c|c|c}
\toprule
\textbf{Model} & \textbf{DBPD} & \textbf{SRD} & \textbf{SMRD} \\
\midrule
ResNet-50 &  79.92 / 21 / 0.3  &  79.11 / 20.8 / 0.93  &  \textbf{80.61 / 20.3 / 0.23}  \\
\bottomrule
\end{tabular}
\end{table}

\textbf{What is the practical usability of SMRD?} To demonstrate the practical usability of SMRD, we derived simple mathematical approximations of the dropout schedule that closely replicate the oracle SMRD performance (Appendix Section~\ref{sec:math_approx_smrd}). Among these, the logarithmic decay approximation provides the closest match. Empirical results in Table~\ref{tab:smrd_log_decay} show that these approximations still clearly outperform the baselines, confirming their practicality. Moreover, SRD and DBPD also outperform the baselines in most scenarios, further validating their effectiveness and applicability.

\begin{table}[h]
\centering
\caption{Practical Usability of SMRD with Logarithmic Decay Approximation.}
\label{tab:smrd_log_decay}
\begin{tabular}{l|l|c|c}
\toprule
\textbf{Model} & \textbf{Variant} & \textbf{Accuracy} & \textbf{Effective Epochs} \\
\midrule
ResNet-50 & SMRD (oracle) &  81.54  &  33  \\
ResNet-50 & SMRD (log decay) &  81.11  &  37  \\
\midrule
EfficientNet-B0 & SMRD (oracle) &  84.45  &  34  \\
EfficientNet-B0 & SMRD (log decay) &  83.92  &  36  \\
\midrule
MobileNet-v2 & SMRD (oracle) &  74.73  &  36  \\
MobileNet-v2 & SMRD (log decay) &  74.36  &  38  \\
\bottomrule
\end{tabular}
\end{table}

\textbf{What is the wall-clock time using PDD?} While our primary metric, Effective Epochs (EE), enables hardware-independent comparison, we also provide several wall-clock time savings estimates in Table~\ref{tab:wall_clock_time} for reference. These experiments were conducted on a single NVIDIA RTX 4060 GPU with 8 GB of RAM.

\begin{table}[h]
\centering
\caption{Wall clock time savings estimates.}
\label{tab:wall_clock_time}
\begin{tabular}{l|c|c|c|c}
\toprule
\textbf{Model} & \textbf{Baseline} & \textbf{DBPD (0.3)} & \textbf{SRD (0.95)} & \textbf{SMRD (0.3)} \\
\midrule
EfficientNet-B0 (CIFAR-100) &  15.74  hr &  7.89  hr &  6.98  hr &  9.95  hr \\
MobileNet-v2 (CIFAR-100) &  12.62  hr &  7.11  hr &  6.67  hr &  9.51  hr \\
\bottomrule
\end{tabular}
\end{table}

\section{Number of samples required for SRD}
In our setting, we use a constant decay $\gamma$ that samples at the same rate from the dataset iteratively. This could be formalized as : 
\begin{equation}
    f(e) = \gamma^e \times N
\end{equation}
where $e$ is the given epoch and $N$ is the number of samples in the dataset. Since we include all the samples as part of our last (revision) epoch $E$, this would be a piecewise function: 
\begin{equation}
f(e) =
    \begin{cases}
        \gamma^e \times N & \text{if } 1 \le e \le E-1\\
        N & \text{if } e = E
    \end{cases}
\end{equation}

\section{Number of samples required for SMRD}
For our SMRD setting, the number of samples chosen is equal to the number of samples that are above the set threshold for epochs $1\le e<E$ where $E$ is the total number of epochs:
\begin{equation}
    f(e) = \#\{x\in \mathcal{D} \mid (p_e(x_i)) < \tau\} \; \forall i=1,....N
\end{equation}  
We know that as the number of epochs progresses, the number of samples gradually reduce. Assuming the model's prediction probabilities $p_e(x)$ follows a beta distribution : $p_e(x) \sim Beta(\alpha_e, \beta)$. So, at each epoch the number of training samples would be:
\begin{equation}
    f(e) = N \times Pr(p_e(x) < \tau) = N \times F_e(\tau; \alpha_e, \beta)
\end{equation}
where $F_e(\tau; \alpha_e, \beta)$ is the cumulative distribution function (CDF) of the Beta distribution.
%%%%%%%%%%%%%%%%%%%%%%%%%%%%%%%%%%%%%%%%%%%%%%%%%%%%%%%%%%%%

\section{MAE pre-training}

In the main paper, we report results by pre-training MAE using SRD 0.98 and this effectively resulted in a 16 $\times$ cheaper pre-training cost whilst showing extremely competitive fine-tuning performance. We also consider a variant of the DBPD by setting a threshold to the MSE loss of each sample. Specifically, we consider two extremes: a threshold of 0.95 which results in most samples being dropped and a threshold of 0.05 which would drop a sample if the MSE is less than 0.05, which helps keep more samples in the pre-training loop. We do a varying amount of revision epochs. For this case, we only perform downstream image classification on ImageNet and \textbf{only for 50 epochs}. Results can be seen in Table~\ref{tab:mae_dbpd}. `No MAE' refers to using a plain ViT-B backbone from scratch (these typically need fine-tuning for over 300 epochs). `DBPD 0.95' ends up stopping pre-training as early as 600 epochs with no samples remaining for the threshold. This is expected as an MSE of 0.95 is very high. Setting such a high threshold naturally results in a very low EE of just 0.69 as most samples are dropped. This also results in poor representations. The result with 1 revision epoch is reflecting that. Matching results of training ViT-B from scratch. Using more revision epochs in this case basically means using MAE for said number of epochs. Using a threshold of 0.05 however is much lower and this results in a much higher EE. However, this also gives us a much higher accuracy, even outperforming the full MAE. 

\begin{table}[t]
\centering
\caption{Accuracy (\%) for ImageNet fine-tuning. `PT EE' corresponds to Pre-training Effective Epochs.}
\begin{tabular}{l|c|c}
\toprule
\textbf{Model } & \textbf{PT EE} & \textbf{Accuracy} \\
\midrule
No MAE & - &  58.61 \\
Full MAE & 800 & 80.26 \\
SRD 0.98 & 49.95 & 77.89 \\
DBPD 0.95 + Revision 1 & 1.69 & 58.05  \\
DBPD 0.95 + Revision 200 & 200.69 & 70.90\\
DBPD 0.05 + Revision 1 & 743.02 & 80.56 \\
DBPD 0.05 + Revision 50 & 792.02 & \textbf{80.65} \\
\bottomrule
\end{tabular}

\label{tab:mae_dbpd}
\end{table}

\section{Estimation of Effective Epochs (EE) for Comparative Methods}

Several prior methods in dataset-level optimization employ data dropout or resampling strategies but do not directly report the number of times the network effectively sees the data during training. To allow a meaningful comparison, we estimate the \textit{Effective Epochs (EE)}---defined as the cumulative number of data sample passes relative to the full dataset size---for both CIFAR-10 and CIFAR-100.

\subsection{DataDropout on CIFAR-10 and CIFAR-100}

The DataDropout method reports removing 1,283 ``unfavourable'' samples after the first epoch. Assuming 30 training iterations for CIFAR-10 and 200 for CIFAR-100, the EE is calculated as follows:

\textbf{CIFAR-10}:
\[
\text{EE}_{\text{CIFAR-10}} = \frac{1 \cdot 50{,}000 + (30 - 1) \cdot (50{,}000 - 1{,}283)}{50{,}000}
= \frac{50{,}000 + 29 \cdot 48{,}717}{50{,}000} \approx 29.26
\]

\textbf{CIFAR-100}:
\[
\text{EE}_{\text{CIFAR-100}} = \frac{1 \cdot 50{,}000 + (200 - 1) \cdot (50{,}000 - 1{,}283)}{50{,}000}
= \frac{50{,}000 + 199 \cdot 48{,}717}{50{,}000} \approx 195.61
\]

\subsection{Importance Sampling on CIFAR-10 and CIFAR-100}

The Importance Sampling (IS) method dynamically resamples the entire dataset at each iteration. Thus, the Effective Epochs are equal to the total number of training epochs used.

\textbf{CIFAR-10}:
\[
\text{EE}_{\text{CIFAR-10}} = 30
\]

\textbf{CIFAR-100}:
\[
\text{EE}_{\text{CIFAR-100}} = 200
\]

Table~\ref{tab:ee-summary} summarizes the reported accuracy and estimated Effective Epochs for each method and dataset.

\begin{table}[h]
\centering
\caption{Reported estimated Effective Epochs (EE) for DataDropout and Importance Sampling (IS) on CIFAR-10 and CIFAR-100.}
\label{tab:ee-summary}
\begin{tabular}{lccc}
\toprule
\textbf{Method} & \textbf{Dataset} & \textbf{EE (Estimated)} \\
\midrule
Baseline & CIFAR-10  & 30 \\
DataDropout & CIFAR-10  & 29.2 \\
IS          & CIFAR-10  & 30 \\
Ours & CIFAR-10  & \textbf{6.98} \\
\midrule
Baseline & CIFAR-100 & 200 \\
DataDropout & CIFAR-100  & 195.61 \\
IS          & CIFAR-100  & 200 \\
Ours & CIFAR-100  & \textbf{28.6} \\
\bottomrule
\end{tabular}
\end{table}

Directly comparing accuracy is hard as DataDropOut~\cite{wang2018data} and IS~\cite{katharopoulos2018not} use different backbones, but our estimations show that the effective epochs saved are extremely marginal whilst their reported improvement over the baseline used is marginal compared to us.

%\section{Test Accuracy vs Epochs curves}

\section{Multiple Runs for testing robustness to random results}
To verify the robustness of our results, we check the standard deviation over five trials as reported in table \ref{tab:5trials}. From the results, we can notice a minimal standard deviation of $\approx 0.2515$ with a mean accuracy of $88.058\%$. We run EfficientNet-B0 on the CIFAR10 dataset. We notice similar results across different models and datasets showing the robustness of the approach.

\begin{table}[h]
\centering
\caption{Accuracy (\%) and Effective Epochs (EE) for five trials done using EfficientNet-B0 with SMRD 0.3 variant on CIFAR10 dataset.}
\begin{tabular}{l|c|c}
\toprule
\textbf{Trial} & \textbf{EE} & \textbf{Accuracy} \\
\midrule
Trial 1 & 6.65 & 88.44\\
Trial 2 & 6.71 & 87.74 \\
Trial 3 & 6.69 & 88.00 \\
Trial 4 & 6.67 & 88.02\\
Trial 5 & 6.74 & 88.09\\
\bottomrule
\end{tabular}
\label{tab:5trials}
\end{table}

\end{document}